\documentclass[11pt]{article}

\usepackage[utf8]{inputenc}
\usepackage[T1]{fontenc}
\usepackage[margin=1in]{geometry}
\usepackage{amsmath,amssymb,amsthm}
\usepackage{booktabs}
\usepackage{graphicx}
\usepackage{microtype}
\usepackage[hidelinks]{hyperref}
\usepackage{float}
\usepackage{multirow}

\newtheorem{lemma}{Lemma}
\newtheorem{definition}{Definition}
\newtheorem{law}{Empirical Law}

\title{\textbf{The Anatomy of a Truth Direction}\\[2pt]
\large Knowledge-Dependent Dimensionality, a Relational Law,\\
and a Shared Category Geometry in Small Language Models}

\author{Francesco Karim Vicidomini\\
\small Independent researcher\\}

\date{July 2026}

\begin{document}
\maketitle

\begin{abstract}
B\"urger et al.\ (2024) demonstrated that truth representations in large language models are universal across statement polarity but reside within a multidimensional subspace.
The truth value of a statement is linearly readable from a residual stream of language model, but it is not clear how much of that representation fits on a single direction, which component builds it, or what it is made of. We conducted a study based on these questions, with one
instrument: a training-free axis, the dominant direction of the singular value decomposition (SVD) of hidden-state differences over true/false minimal pairs, identified without labels up to one global sign.
Extensive evaluation across 14 models from 6 diverse architectural families (including MoE), read and extract at cost $O(d)$ per token.
We close with a pre-registered prediction on whether the arrangement extends to categories whose truth is computed rather than retrieved.
\end{abstract}

\section{Introduction}

Large language models produce false statements with the same fluency as
true ones, and a growing body of work shows that the truth value of a
statement is linearly readable from the residual
stream~\cite{burns2023,marks2024,li2023,zou2023}. B\"urger, Hamprecht and
Nadler~\cite{burger2024} sharpened this picture in a way that motivates
everything below: truth is \emph{universal} across statement polarity, but
it is not a single direction; probes trained on affirmative statements fail
on negated ones, and the representation lives in a two-dimensional subspace
with a dedicated polarity direction. If the truth representation has more
structure than one direction, the questions follow naturally: How much of
truth fits on a single direction, and when does that reading fail? Which
architectural component builds the direction, and which degrades it? And
what is the direction made of, when the material spans many kinds of facts?

This manuscript answers the three questions, with one
instrument and one discipline. The instrument is a training-free axis: the
direction of the SVD of hidden-state differences, true/false
minimal pairs, identified without labels up to sign
(Lemma~\ref{lem:labels}), read at cost $O(d)$ per token, and evaluated
held-out with a permutation null that prices the optimism of layer
selection. The discipline is falsification: predictions pre-registered,
negative results are reported with the same care as positive
ones, and two corrections of our own published interpretations are narrated
in the text as results, because a claim that has survived its author's own best
attack is worth more than one that has never been attacked.

\textbf{Part~I establishes that the signal is real but bounded, and that its
dimensionality tracks what the model knows: the gap between a full linear probe and the single axis widens with the fraction of facts the model cannot produce.
Part~II locates the mechanics. Across fourteen models, six families and five
architectural placements, attention leads the FFN at the readability peak and
the FFN's contribution turns against the axis one block later. A directional
intervention separates rotation from erosion: removing the component parallel
to the axis recovers readability in $17$ of $18$ block-interventions
(sign test $p=7.3\times10^{-5}$), while removing a random direction of the same
construction changes nothing. Inside the FFN the anti-truth write is carried by
the value stream, not by the gate. Part~III measures how the axis is composed.
Per-category axes are arranged consistently across models: all $91$ pairwise
agreements are positive, with a mean of $0.798$ after correcting for each
model's own measured reliability, and the arrangement does not exist at all at
the embedding layer, where the $528$ off-diagonal cosines are identical.}

Two pre-registered predictions failed and are reported as such. The hypothesis
that the shared arrangement reflects shared distillation lineage is not
supported: two models trained from scratch without a teacher rank above both
distilled families. And the tug-of-war between attention and the FFN after the
peak, previously recorded as a trait of scale on four models, is not ordered by
parameter count once ten are measured.

Throughout, predictions are written before the runs, correctness gates halt
the pipeline rather than warn, and the sentence construction, the sign gauge
and the knowledge threshold are declared rather than assumed. Code, "dictionaries-bundles" and the exact command for every table are released.

\part{The Axis on One Model}
\addcontentsline{toc}{part}{Part I}

\section{Method}
\label{sec:method}

\subsection{Extraction}
\label{sec:extraction}

Informally: a transformer reads a sentence and, at every layer, updates an
internal vector summary of the text so far; we record that summary at the
position of the last token, where the sentence is complete, at every layer,
and study the geometry of these vectors.

\subsection{Sentence construction, and what it decides}
\label{sec:convention}

Every state is read at the last token of a sentence built as prompt, space,
target, followed by a period. The dataset does not fix what follows the
target, and the choice is not cosmetic: it decides what the measurement can
see.

With the period, the last token is identical in the two sentences of a pair,
so the identity of the target does not enter the pair difference and whatever
separates the two states was computed and transported to a position that does
not contain the answer. Without it, the last token \emph{is} the target, and
the difference mixes the model's judgement with the identity of two different
words.

We measured both. Without the period, held-out AUC at layer $0$, the
embeddings, before any block has computed anything, is $0.534$ on average
across six models (range $0.480$--$0.549$) rather than $0.500$; and the gap
between a full linear probe and the single axis, the quantity that carries the
dimensionality claim of Sec.~\ref{sec:dim}, falls from a mean of $0.094$ to
$0.016$, reaching exactly zero on one model and a negative value on another. A
truth signal that is one-dimensional in every model regardless of architecture
is not a property of the models: it is the target token, which occupies a
single direction of embedding space.

The period is therefore the condition that makes the ``curve starts at
$0.500$ at layer $0$'' control meaningful, and the condition under which the
dimensionality claim is testable at all.

CounterFact was built for activation patching on the subject token, not for
reading a verdict at the final position. Using it as we do requires a sentence
construction the dataset does not impose, and the consequences above are the
price of that choice.

\subsection{Data: minimal pairs}
\label{sec:data}

A \emph{minimal pair} is a pair of sentences with the same subject and the
same template that differ in exactly one content word, such that one sentence
is true and the other false: ``The Sun is a star.''\ vs.\ ``The Sun is a
planet.''\ Both sentences are affirmative. Within a pair, the difference of
hidden states cancels the shared topic and surface form and leaves the truth
signal.

The \emph{curated pairs} are a built-in set of $45$
pairs of common facts (capitals, chemical symbols, authorship, currencies)
that the model demonstrably reproduces in open-ended completion; they play the role of a known-facts control.
The external datasets are CounterFact~\cite{meng2022} and
TruthfulQA~\cite{truthfulqa} ($250$ pairs, are default in code), which contain many relations a $1.5$B model plausibly
does not know.

To quantify this, we define the \emph{know-rate}: the fraction of facts the
model reproduces correctly in open-ended greedy completion. Over $200$
CounterFact prompts, Qwen2.5-1.5B has a know-rate of $32.0\%$ ($64/200$),
which grounds the ``plausibly does not know'' reading; the know-rates of the
other models appear in Part~II.

Precisely: a prompt counts as known when the greedy completion
(\texttt{do\_sample=False}, at most $8$ new tokens) either begins with the
true target or contains it preceded by a space, case-insensitively
(\texttt{behav\_check.py}). Three properties of this estimator matter for
what follows. It is a \emph{lower bound}: a model may know a fact and
still continue differently within eight tokens. It is \emph{greedy}: no
sampling, so the estimate is deterministic given the seed that selects the
prompts. And it inherits tokenizer and completion-style differences
between families, so cross-family comparisons carry a systematic bias
component, while within a family the bias is shared and cancels. This is
why the dimensionality relation is stated as within-family monotone with a
cross-family offset, and why the ordering Llama-3.2-1B ($36\%$) above
Qwen2.5-1.5B ($32.0\%$) is neither exotic nor load-bearing. Parameter
count is not knowledge: the two models differ in training data, tokenizer,
and completion style, and no claim below compares know-rates across
families.

\paragraph{Why a flipped predicate and not a negation.} The false counterpart
is built by flipping a content word (``planet'' for ``star''), not by
inserting ``not''. The choice is deliberate, for three reasons. First,
negation is not a neutral edit: it activates a dedicated polarity direction in
the representation~\cite{burger2024}, so a negated counterpart would entangle
the truth signal with the polarity signal. Our design keeps both sentences
affirmative and isolates truth; polarity is then studied separately and
deliberately in the $2{\times}2$ design of Sec.~\ref{sec:poldesign}. Second,
inserting ``not'' changes the token count and shifts all subsequent positions,
weakening the token-level matching that makes the intra-pair difference
cancel shared structure. Third, the two designs are complementary rather than
competing: the negated forms are exactly the NT/NF cells of the polarity
design, so the alternative is not ignored but treated as its own object of
study.

\subsection{The axis, and where labels enter}
\label{sec:axis}

We form the intra-pair difference matrix $D$ with rows
$h_{\text{true}} - h_{\text{false}}$ at a given layer, and take its SVD,
\begin{equation}
D = U S V^{\top}, \qquad v_{1} = V_{1},
\end{equation}
where $v_1$, the top right-singular vector, is the dominant direction of the
differences. The role of supervision in this construction can be stated
exactly.

\begin{lemma}[Label-free identification of the axis, up to sign]
\label{lem:labels}
Let each pair $i$ contribute a row $d_i^{\top}$ to $D$, where
$d_i = s_i\,(h_i^{A} - h_i^{B}) \in \mathbb{R}^{d}$ is a column vector,
$h_i^{A}, h_i^{B}$ are the two hidden states of the pair, and
$s_i \in \{\pm 1\}$ encodes which sentence of the pair is designated true.
Then $D^{\top} D = \sum_i d_i d_i^{\top}$ does not depend on $(s_i)$; hence
the right singular vectors of $D$, and in particular the span of $v_1$, are
invariant to the within-pair labels.
\end{lemma}

One implicit choice is declared here rather than discovered by a referee:
the SVD is applied to the raw difference rows, so $v_1$ is a
\emph{norm-weighted} principal direction, with pair $i$ contributing
weight proportional to $\lVert d_i \rVert^{2}$. If difference norms
correlate with sentence length or with the state norm at the reading
position, long or high-norm pairs dominate the fit. We conducted a test using some of the models, at their peaks
(\texttt{axis\_norm\_check.py}): the row-normalized refit agrees with the
raw axis at $|\cos| = 0.984$, $0.980$, $0.974$, $0.966$ (Qwen2.5-1.5B,
Qwen2.5-3B, Llama-3.2-1B, Llama-3.2-3B), the held-out AUC changes by at
most $+0.016$, and is in fact slightly higher under normalization on every
model, so the axis does not ride on high-norm pairs; difference norms
correlate only weakly, and negatively, with sentence length ($r$ between
$-0.12$ and $-0.23$). The axis is not a norm-weighting artifact.

\paragraph{What label-free does and does not mean.} Three levels of
supervision must be kept apart. (i) \emph{Dataset construction} is
supervised: writing a minimal pair requires knowing which completion is
factually correct, and no method that starts from such pairs can call
itself unsupervised at this level. (ii) \emph{Axis estimation} consumes
only the contrast structure: by Lemma~\ref{lem:labels}, the span of $v_1$
is invariant to which element of each pair is designated true, so an
adversary who scrambled every within-pair label would hand the algorithm
the same axis. The information the estimator uses is ``these two
statements differ in truth value'', not ``this one is true''.
(iii) \emph{Orientation and evaluation} are supervised: one global sign
bit, and held-out labels for scoring. Accordingly, this manuscript says
\emph{training-free} of the pipeline, \emph{label-free up to sign, given
the paired design} of the estimator, and reserves \emph{unsupervised} for
the polarity recovery of Part~I, where the recovered factor was never
encoded in any label handed to the algorithm.

Labels enter the pipeline at exactly two points. First, a single global
orientation bit (``the true side is positive''), fit on training folds,
fixes the sign of $v_1$. Second, the evaluation itself: AUC is defined with
respect to labels. Everything upstream of these two points, the pairing, the
differences, and the axis, uses no labels. We therefore describe the axis as
\emph{training-free} (no optimization of any kind) and \emph{label-free up to
one sign bit} (Lemma~\ref{lem:labels}), and we reserve the word
``unsupervised'' for the polarity recovery of Sec.~\ref{sec:polarity}, where
the recovered structure (polarity) is never labeled at all. The lemma also
explains, in advance, the behavior of the permutation null below: permuting
which sentence of a pair is called true leaves the span of the axis
unchanged and destroys only the orientation and the test labels, which is why
the null concentrates near chance.

\subsection{A falsifiable evaluation protocol}
\label{sec:protocol}

The central methodological point is that the axis, its calibration, the
orientation bit, \emph{and} the choice of peak layer (the layer whose held-out AUC is maximal, chosen on training folds) must be fit on training
data only. We use $k$-fold cross-validation \emph{over pairs}: a pair is
never split across folds, which would leak the shared topic.

Part of the apparent accuracy of any layer-scanned pipeline is selection: the
best of $29$ layers looks good by chance. The \emph{permutation null}
quantifies exactly that contribution. Let $A_{\mathrm{obs}}$ be the observed
maximum held-out AUC over layers (AUC: the area under the ROC curve; equivalently, the probability that a
randomly chosen true statement scores above a randomly chosen false one,
ties counted as one half. For $b = 1, \dots, B$ we randomly swap
true/false \emph{within} each pair (the labels become noise while the pair
structure is preserved), rerun the identical pipeline including the layer
scan, and record the maximum held-out AUC $A_b$. The estimated $p$-value is
the standard conservative estimator
\begin{equation}
\hat{p} \;=\; \frac{1 + \#\{\,b : A_b \ge A_{\mathrm{obs}}\,\}}{B + 1}.
\end{equation}
With $B = 200$ and no null value reaching the observed statistic,
$\hat{p} = 1/201 \approx 0.005$, the value reported in
Table~\ref{tab:main}. As a reference point we also fit an
$\ell_2$-regularized logistic probe on the raw $d$-dimensional states, held
out over the same pair folds.

\subsection{Polarity design}
\label{sec:poldesign}

To test whether a genuine second dimension exists, we use a $2{\times}2$
design: each fact in four forms, affirmative-true (AT), affirmative-false
(AF), negated-true (NT), negated-false (NF), giving the XOR structure
of~\cite{burger2024}. We compute the affirmative and negated truth directions
$t_A = \overline{AT} - \overline{AF}$, $t_N = \overline{NT} - \overline{NF}$,
the general direction $t_G \propto t_A + t_N$, and the supervised polarity
direction $t_P \propto t_A - t_N$. For the recovery test, we run the
difference SVD on minimal pairs that \emph{mix} the two polarities, so the
SVD never sees the polarity label, and compare its leading directions to the
supervised $t_P$.

\section{Results I: signal, dimensionality, polarity}
\label{sec:results}

All numbers are on Qwen2.5-1.5B with last-token pooling; AUC is out-of-fold
unless stated.

\subsection{A real but bounded truth signal}
\label{sec:signal}

Table~\ref{tab:main} reports the held-out evaluation at the peak layer (layer
$16$). On the curated pairs the training-free axis reaches AUC $0.938$ with
permutation $\hat{p} = 0.005$ (null mean $0.588$, $95$th percentile $0.664$):
the geometric axis aligns with truth well beyond what layer selection alone
can manufacture. The signal is therefore real. It is also \emph{bounded}: a
full linear probe on the same raw states reaches $0.989$. The probe sees
information the single axis discards.

\begin{table}[htbp]
\centering
\caption{Held-out AUC ($5$-fold CV over pairs) at layer $16$, Qwen2.5-1.5B.
``1D'' is the real axis; ``MAG'' the magnitude-gated polar score; ``PHASE''
the angular-deviation score; ``2D'' a logistic on
$(\mathrm{Re},\mathrm{Im})$; ``Probe'' an $\ell_2$ logistic on raw states.
The last column is the label-permutation $\hat{p}$ for the 1D axis.}
\label{tab:main}
\begin{tabular}{lccccccc}
\toprule
Dataset & pairs & 1D & MAG & PHASE & 2D & Probe & $\hat{p}$ \\
\midrule
Curated pairs                    & $45$  & $0.938$ & $0.936$ & $0.928$ & $0.931$ & $0.989$ & $0.005$ \\
CounterFact only                 & $250$ & $0.792$ & $0.790$ & $0.771$ & $0.795$ & $0.854$ & $0.005$ \\
CounterFact / TruthfulQA mix     & $250$ & $0.638$ & $0.640$ & $0.676$ & $0.651$ & $0.841$ & $0.005$ \\
\bottomrule
\end{tabular}
\end{table}

\subsection{The gap scales with model knowledge, and with heterogeneity}
\label{sec:dim}

The rows of Table~\ref{tab:main} differ not only in difficulty but in the
\emph{size of the probe gap}, and the three rows separate two distinct
effects.

The first effect is knowledge. On curated facts the model demonstrably knows,
the axis ($0.938$) trails the full probe ($0.989$) by $0.051$. On CounterFact
alone, where the measured know-rate is $32.0\%$ (Sec.~\ref{sec:data}) and
many relations are obscure (for example, the mother tongue of a minor
historical figure), the separability of the axis falls to $0.792$ and the gap
widens to $0.062$ (probe $0.854$). Less knowledge, lower separability, and a
modestly larger gap.

The second effect is heterogeneity. On the CounterFact/TruthfulQA mix the gap
jumps to $0.203$ (axis $0.638$, probe $0.841$), far more than either
ingredient alone would suggest. A single direction fit across heterogeneous
material must serve all of it at once, while the full probe is free to
combine directions; the excess gap on the mix therefore measures how much the
truth direction itself varies with the material. Subsequent work confirms
this reading independently (Part~III): the
per-category truth axes form a structured, semantically signed geometry, and
forcing one axis onto a mixture of categories is exactly what the mix row
pays for.

We read the knowledge effect as a statement about \emph{dimensionality}. When
the model knows a fact, its truth content is concentrated and a single
direction captures most of it. When the model does not know a fact, the
truth-relevant signal is weaker and more diffuse, and one direction recovers
less of it. The dimensionality of the truth representation is thus not fixed
but \emph{knowledge-dependent}; the mix row adds the complementary statement
that it is also \emph{category-dependent}. This joint framing is, to our
knowledge, stated in these terms by neither the linear-probe nor the
dictionary-learning literature, and it is the contribution we consider most
novel.

\subsection{The complex plane does not help; ``phase'' is the sign flip}
\label{sec:phase}

On single-polarity data the second dimension adds nothing: at layer $16$ the
2D logistic ($0.931$) does not exceed the 1D axis ($0.938$) on curated pairs,
and on CounterFact it gains $+0.003$ ($0.795$ vs.\ $0.792$). This is
structural. Minimal pairs cancel the topic, so the differences concentrate
along one direction; the SVD places the entire signal in $v_1$, and $v_2$ is
residual variance, noise rather than a second truth factor. A controlled
synthetic check confirms that the apparent ``rotation'' of falsehoods is an
artifact: with a pure sign flip and zero genuine rotation, an angular (phase)
score still separates classes, because a state at $\mathrm{Re} < 0,\
\mathrm{Im} \approx 0$ has phase $\approx \pi$ by definition. The earlier
intuition of truth-as-phase has no operational content here: it reduces to
the sign of the real axis re-expressed in polar coordinates. We retract that
framing.

\subsection{Recovery of the polarity direction without polarity labels}
\label{sec:polarity}

A genuine second dimension does exist, but only when the data span both
polarities, and it is the polarity direction of~\cite{burger2024}, not a
phase. Table~\ref{tab:polarity} shows the diagnostic: an axis fit on
affirmative statements classifies affirmative held-out facts at $0.981$ but
\emph{anti-classifies} negated ones at $0.079$, a near-total flip, while the
general direction $t_G$ recovers the negated facts at $0.954$. The truth
direction genuinely rotates with polarity ($\cos(t_A, t_N) = -0.415$).

The methodological result is the recovery. When we run the difference SVD on
\emph{mixed-polarity} minimal pairs, blind to the negation label, its leading
direction aligns with the supervised polarity direction $t_P$ at cosine
$0.959$ (layer $15$), and $t_P$ lands in the first principal component, while
alignment with $t_G$ is only $0.157$. In other words, when both polarities
are present, the dominant axis of the differences \emph{is} polarity, and the
SVD finds it without supervision; by Lemma~\ref{lem:labels} the recovery
cannot have used the labels even implicitly. Where~\cite{burger2024}
constructs $t_P$ by separating affirmative from negated statements by hand,
the same direction emerges here from geometry alone.

\begin{table}[htbp]
\centering
\caption{Polarity analysis (leave-facts-out CV), layer $14$ chosen on
affirmative data; and unsupervised $t_P$ recovery at layer $15$. AUC below
$0.5$ indicates a sign flip.}
\label{tab:polarity}
\begin{tabular}{lc}
\toprule
Quantity & Value \\
\midrule
$\cos(t_A, t_N)$ & $-0.415$ \\
Affirmative axis on affirmative (held-out AUC) & $0.981$ \\
Affirmative axis on \emph{negated} (held-out AUC) & $0.079$ \\
General direction $t_G$ on negated (held-out AUC) & $0.954$ \\
\midrule
SVD vs.\ supervised $t_P$: $\max_k |\cos(\mathrm{SVD}_k, t_P)|$ & $0.959$ \\
\quad (principal component carrying $t_P$) & PC\,$1$ \\
$\cos(\mathrm{SVD}_1, t_G)$ & $0.157$ \\
\bottomrule
\end{tabular}
\end{table}

\section{Robustness: seven falsification experiments}
\label{sec:robust}

The preceding section established a positive claim. We now stress-test it. The
claim under test is:

\begin{quote}
\emph{In Qwen2.5-1.5B, the truth value of a statement is carried by a single
direction, the position axis obtained from the SVD of paired true/false
differences, on facts the model knows; the second and higher dimensions add no
separation.}
\end{quote}

A claim of one-dimensionality is falsifiable in a specific way: exhibit a
second, third, or depth-distributed coordinate that separates true from false
better than the single axis. We construct seven and test each.

\paragraph{Two kinds of test, kept distinct.} \emph{Controlled experiments}
(1--4) produce a held-out AUC under $k$-fold cross-validation over pairs and a
label-permutation null that prices selection bias; a result counts only if it
clears that null. \emph{Visual observations} (5--7) do not classify: they
project the states and measure one descriptive statistic, so the conclusion
follows from a number attached to a picture rather than from a fitted model.
Throughout, the model is frozen Qwen2.5-1.5B, last-token pooling, peak layer
$16$ unless stated.

Table~\ref{tab:seven} carries all seven with their numbers. Two of them
changed something and are discussed below; the remaining five are read from
the table, and their details are reported below.

\begin{table}[htbp]
\centering
\caption{Seven experiments attempting to falsify the one-dimensional truth
axis. ``Exp.'' = controlled experiment (held-out CV + permutation null);
``Obs.'' = visual observation (descriptive separation statistic). None
overturns the claim.}
\label{tab:seven}
\small
\begin{tabular}{p{0.5cm}p{4.6cm}p{1.1cm}p{4.6cm}p{1.4cm}}
\toprule
\# & Richer structure hypothesized & Type & Result (real numbers) & Verdict \\
\midrule
1 & A complex/phase second dimension (truth as rotation) & Exp. & PHASE AUC $0.928$ vs.\ 1D $0.938$ (curated); reduces to the real-axis sign flip & falsified \\
2 & Truth is global across depth (fixed-axis signature) & Exp. & global $0.923$ vs.\ local $0.938$ ($-0.015$), $\hat{p} = 0.0099$; persistence gap $+0.04$ & falsified \\
3 & Truth is dynamic, a ``butterfly effect'' in the unsaturated shade, on hard data & Exp. & global $0.642$ vs.\ local $0.627$ ($+0.015$), $\hat{p} = 0.0099$; commitment true $2.83 \to 5.91$ vs.\ false $2.65 \to 5.86$ & falsified \\
4 & Truth is distributed across layer ``departments'' & Exp. & diagnostic curve is a smooth ramp (no steps); global $0.646$ vs.\ local $0.627$ ($+0.019$), $\hat{p} = 0.0099$ & falsified \\
5 & The argument $\theta$ (phase) carries truth & Obs. & $\theta$ AUC $0.504$ / $0.584$ / $0.556$ (CF / TQA / mix) & falsified \\
6 & The energy $|z|$ (commitment) carries truth & Obs. & $|z|$ AUC $0.560$ / $0.558$ / $0.543$ & falsified \\
7 & True/false form distinct 3D clumps & Obs. & centroid ratio $0.55$ / $0.51$ / $0.48$ ($<1$, overlapping); $\mathrm{Re}$ alone $0.795$ / $0.656$ & falsified \\
\bottomrule
\end{tabular}
\end{table}

\paragraph{1. The complex plane, retracted.} The original framing promoted the
second SVD direction to an imaginary axis and read truth as a phase rotation.
If real, a phase- or magnitude-based score should beat the real axis. It does
not: on curated pairs the phase score reaches AUC $0.928$ against $0.938$ for
the real axis. A controlled synthetic check settles the mechanism: with a pure
sign flip and \emph{zero} genuine rotation an angular score still separates,
because a state at $\mathrm{Re} < 0$ has phase $\approx \pi$ by definition.
The rotation is the sign of the real axis re-expressed in polar coordinates.
The complex framing is decorative and we retract it.

\paragraph{2--4. Depth, in three attacks.} The three remaining controlled
experiments test one target from three angles: that truth is distributed
across depth rather than concentrated at one layer. Experiment~2 gives a
classifier the full trajectory of the signed deviation
$\Delta_\ell = \sigma(\mathrm{Re}(z_\ell)) - \tfrac12$ along a fixed axis.
Experiment~3 repeats it on hard data with the unsaturated features
$[\mathrm{Re}, |z|]$, since $\sigma$ compresses large values and could hide a
propagating uncertainty. Experiment~4 gives every layer its own native probe,
on the hypothesis that different bands do different jobs.

All three answer the same way: the depth signature is real
($\hat p = 0.0099$ in each case) and marginal, at $-0.015$, $+0.015$ and
$+0.019$ AUC against the single best layer. And in each case a descriptive
measure rules out the mechanism proposed. The crossing-persistence gap between
false and true is $+0.04$ and flat, so nothing propagates. The commitment
magnitude grows identically for both classes (true $2.83 \to 5.91$, false
$2.65 \to 5.86$), so false statements do not hesitate more than true ones.
And the per-layer separation curve is a smooth ramp to a single peak at layer
$16$, with no plateaus and no steps, so there are no layer departments.
Recent monitoring methods model the evolution of hidden states across depth rather than reading a single layer~\cite{icr2025,hsad2025}. Our depth trajectories are descriptive and consistent with this line.

\begin{figure}[htbp]
\centering

\begin{minipage}{0.49\textwidth}\centering
\includegraphics[width=\textwidth]{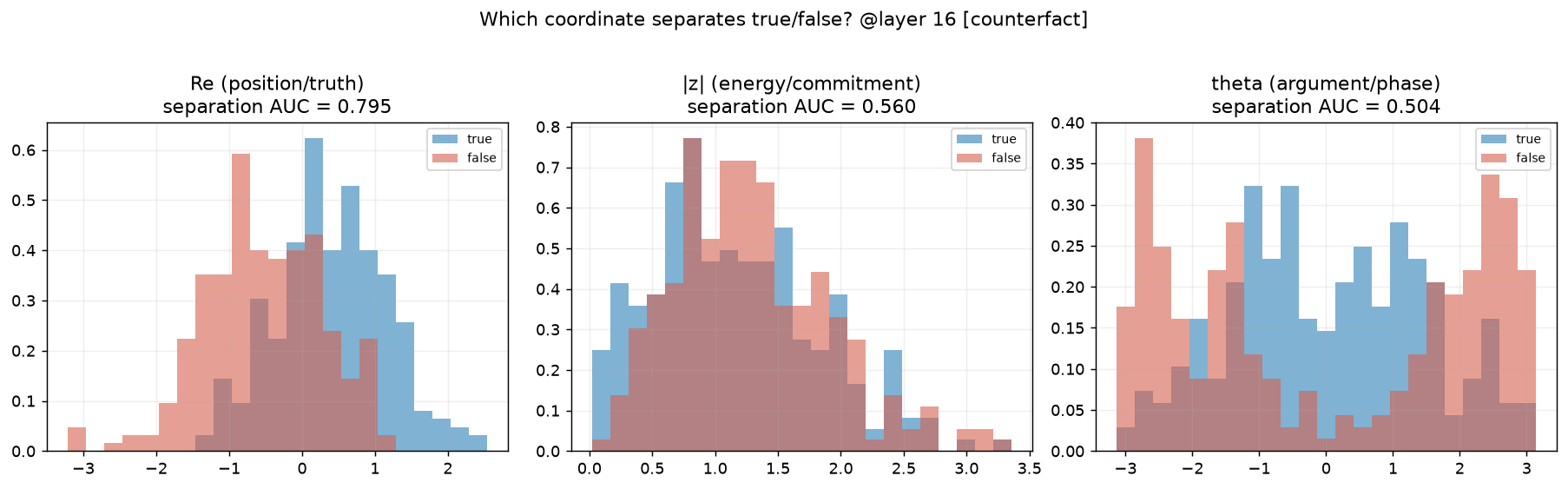}\\[2pt]
{\footnotesize (a) CounterFact: $\mathrm{Re}$ $0.795$, $|z|$ $0.560$, $\theta$ $0.504$}
\end{minipage}\hfill
\begin{minipage}{0.49\textwidth}\centering
\includegraphics[width=\textwidth]{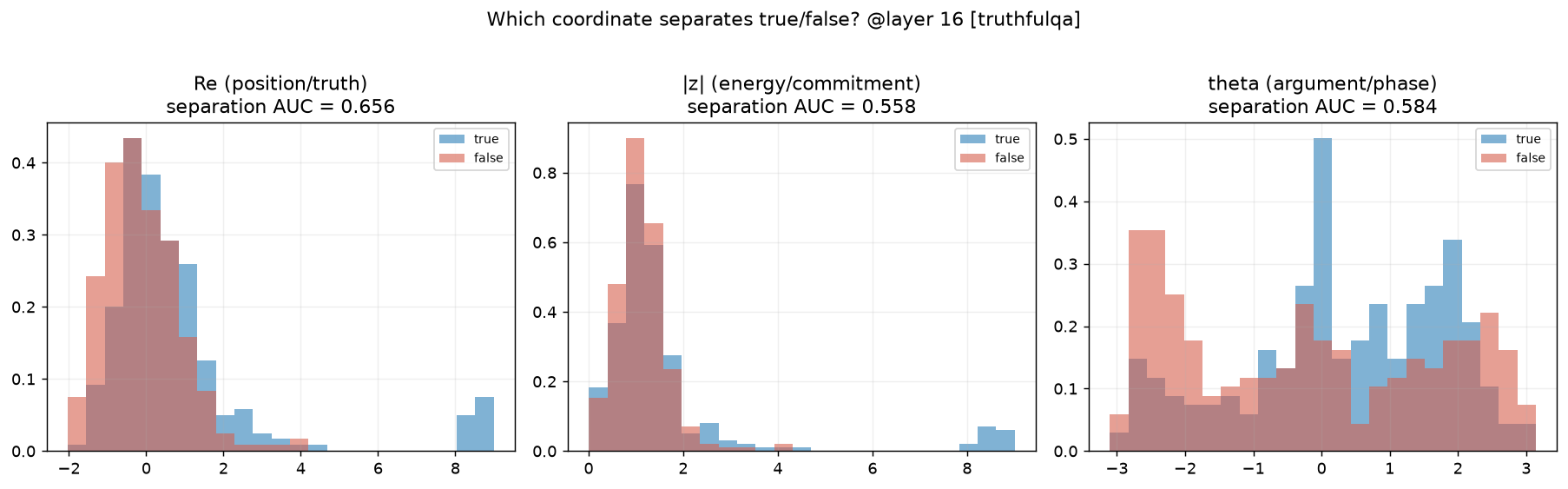}\\[2pt]
{\footnotesize (b) TruthfulQA: $\mathrm{Re}$ $0.656$, $|z|$ $0.558$, $\theta$ $0.584$}
\end{minipage}
\caption{Experiments 5--6 and knowledge scaling, layer $16$. Per-coordinate
separation AUC. On both datasets only the position axis ($\mathrm{Re}$)
separates true from false; energy ($|z|$) and argument ($\theta$) are at
chance. The position separation falls from $0.795$ (CounterFact, often-known
facts) to $0.656$ (TruthfulQA, more ambiguous), while the one-dimensional
structure is unchanged.}
\label{fig:coords}
\end{figure}

\begin{figure}[htbp]
\centering

\begin{minipage}{0.49\textwidth}\centering
\includegraphics[width=\textwidth]{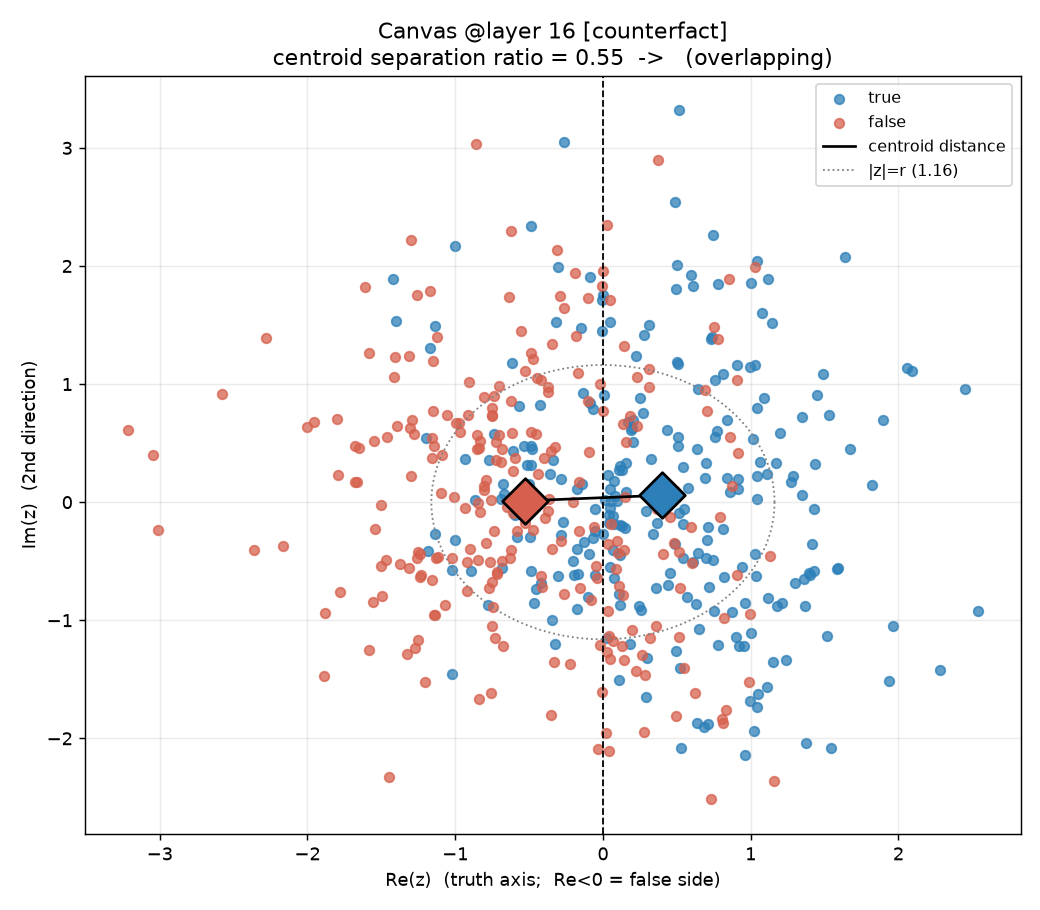}
\end{minipage}\hfill
\begin{minipage}{0.49\textwidth}\centering
\includegraphics[width=\textwidth]{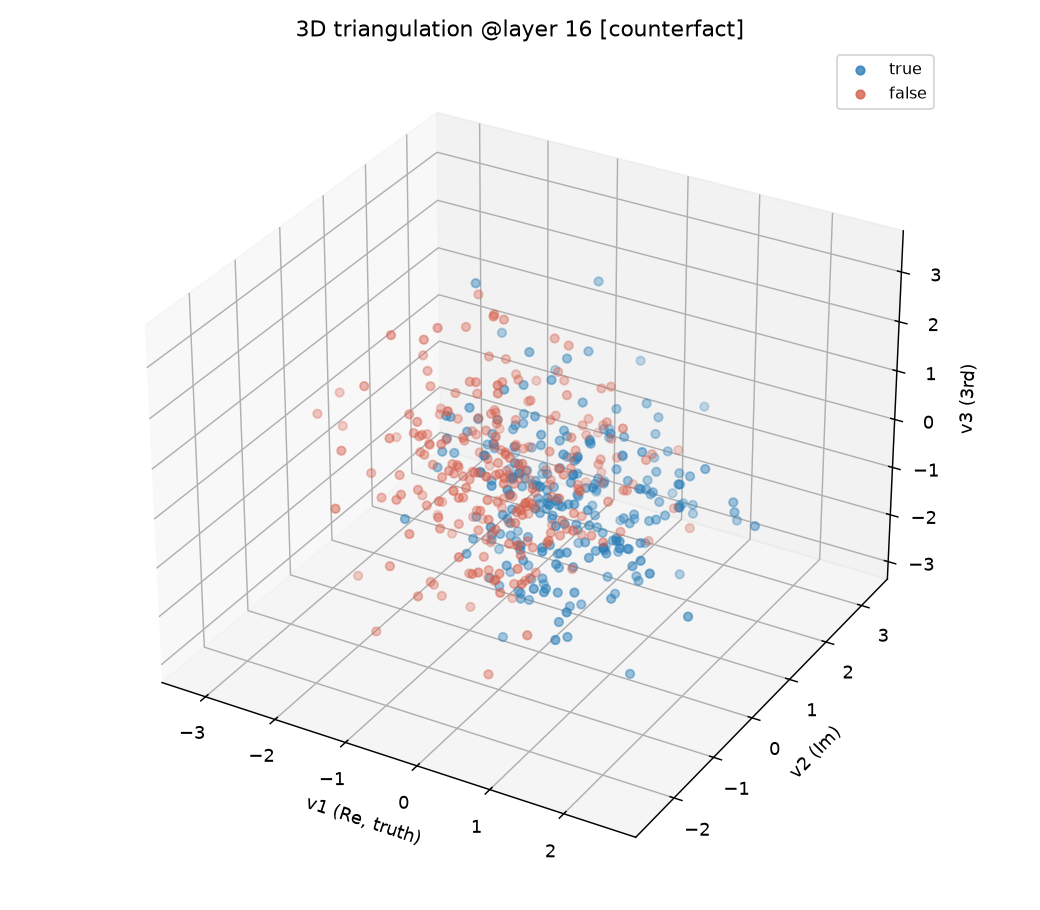}
\end{minipage}
\caption{Experiment 7, CounterFact at layer $16$. Left: the 2D plane; the
true (blue) and false (red) centroids (diamonds) are close relative to the
cloud spread (centroid-separation ratio $0.55$, overlapping). Right: the
three-dimensional $(v_1, v_2, v_3)$ scatter shows a gradient along $v_1$
(position) but no distinct clumps. The separation is confined to the
horizontal position axis; the extra dimensions are noise.}
\label{fig:plane}
\end{figure}

\paragraph{5--7. Energy, argument, and clumping.} Decomposing each state into
position ($\mathrm{Re}$), energy ($|z|$) and argument ($\theta$), only
position carries signal: energy separates at $0.560/0.558/0.543$ and argument
at $0.504/0.584/0.556$ across CounterFact, TruthfulQA and the mix
(Fig.~\ref{fig:coords}). The phase, the coordinate the complex framing made
central, is empty. In three dimensions true and false do not form distinct
clusters: the centroid-separation ratio is $0.55/0.51/0.48$, all below one
(Fig.~\ref{fig:plane}).

That low ratio is \emph{not} evidence against the axis, and the distinction
matters. It is the one-dimensional signal seen through two noise dimensions:
projected onto position alone the same data separate at AUC $0.795$ on
CounterFact. Adding orthogonal dimensions dilutes; it does not enrich.

\subsection{What survives}
\label{sec:survive}

Three statements resist all seven experiments.

\emph{Truth is one-dimensional and lives on the position axis.} Across every
decomposition, only $\mathrm{Re}$ separates; energy, phase, the second and
third SVD directions and the full-depth trajectory add at most $+0.02$ AUC and
usually nothing.

\emph{Separability scales with knowledge.} The position axis separates at AUC
$0.795$ on CounterFact, where the model often knows the fact, at $0.656$ on
the more ambiguous TruthfulQA, and on the diffuse mix the centroids nearly
coincide. The clearer the model's knowledge, the more one-dimensional and
separable the representation.

\emph{On unknown facts the classes overlap.} Where the model does not know,
true and false representations do not form distinct clusters but overlap: a
dense, uniform distribution rather than two clumps. We state this as low class
separability with distribution overlap (centroid ratio $<1$), the measured
counterpart of the knowledge-dependent dimensionality of Sec.~\ref{sec:dim},
and take it up mechanistically in Sec.~\ref{sec:why}.

\section{Why unknown facts overlap: the signature of superposition}
\label{sec:why}

It remains to interpret \emph{why} the classes overlap on unknown facts
rather than simply asking \emph{that} they do. The overlap is not measurement
noise; it is the observable signature of unresolved superposition.

A projection onto a single direction returns a scalar: it collapses the
$d$-dimensional state to one number and discards $d - 1$ dimensions. When the
model knows a fact, its truth content has been allocated a clean, dominant
direction of its own (the residual stream has, in effect, dedicated an axis
to it), and the single projection recovers almost all of it (position AUC
$0.795$, near the full-probe ceiling). When the model does not know a fact,
there is no dedicated direction: the weak truth-relevant signal remains
distributed across many superposed features, and no single axis can pull it
out of the tangle; recovering it is exactly the per-atom expansion of
Sec.~\ref{sec:discussion}, not a cheaper projection.

The consequence is exactly what the visual observations show. On known facts
the truth occupies its own direction, the classes form two regions, and the
centroid ratio is high. On unknown facts the truth is spread across the
superposition, the classes overlap, and the centroid ratio falls below one.
The dense, uniform overlap of true and false on unknown facts is therefore
not a failure of the probe but a direct, visual readout of superposition that
the probe structurally cannot resolve. The probe gaps of
Sec.~\ref{sec:dim} and the low centroid ratio measured here are views of the
same quantity: the truth information that lives off-axis on facts the model
does not know ($0.062$), inflated further when heterogeneous material forces
one axis to serve many truth frames at once ($0.203$). The subsequent work of Part~III begins to resolve that superposed remainder into named, signed components: what this paper can only call
unresolved superposition acquires, there a geometry.

\section{Verify behavior before geometry}
\label{behav}
\paragraph{An informal behavioral check (not a controlled experiment).}
Before building any geometric study of a capability, one should confirm that
the capability exists behaviorally. We ran a small, qualitative probe of the
base model in open-ended completion (greedy decoding, no context carried
between prompts). Two things were apparent on a handful of prompts. First,
the model reproduces the common semantic facts used throughout this paper
(capitals, chemical symbols, authorship, currencies), consistent with the
``facts the model knows'' assumption behind the curated pairs. Second, on
symbolic transitive syllogisms built from nonce words (``a \emph{grok} is
heavier than a \emph{blarn}; a \emph{blarn} is heavier than a \emph{fendu};
therefore $\ldots$''), the base $1.5$B model does \emph{not} perform the
deduction: it deflects into multiple-choice or code scaffolding rather than
completing the transitive conclusion. We stress that this is an informal
observation on a few prompts, not a measured result, and we draw no
quantitative claim from it.

\paragraph{The principle.} Studying the geometry of a capability presupposes
the capability. Measuring the hidden-state geometry of a task the model
cannot perform yields noise, and that noise is easily mistaken for ``the
property is distributed/complex'' when the truthful reading is ``the property
is absent.'' A cheap behavioral feasibility check should therefore precede
any geometric study of \emph{logical} truth; on this model that study is not
feasible, which is itself useful to know in advance. The same principle
applies to transfer: before studying the geometry of truth in a second model,
one should first verify behaviorally that the second model knows the facts in
question. (In the subsequent multi-model work of
Part~II this principle is applied systematically: the
behavioral know-rate is measured before any geometry.)

\paragraph{A hypothesis to test, not a claim.} The way the base model
deflects on uncertain input, falling back on memorized \emph{format}
scaffolding (exam layouts, question/answer templates, code), suggests, but
does not establish, that part of the overlap on unknown facts may carry a
sentence-format component rather than pure superposition noise. This is
consistent with the format confound that makes mixing datasets inadvisable. We flag it as a direction for  future, properly
controlled experiment, not as evidence.

\section{Discussion: what a single axis can and cannot read}
\label{sec:discussion}

It is tempting to view a cheap geometric probe as a low-cost version of
dictionary learning, reading superposition without expanding it. This is
impossible, and the impossibility is a capacity statement rather than an
implementation gap. A projection is an inner product: it collapses a
$d$-dimensional state to a single scalar, discarding $d-1$ dimensions. From
that scalar one cannot reconstruct which features were present, only how much
a chosen direction was. Reading the superposition requires asking the question
once per dictionary atom, which over an overcomplete dictionary of size $m$ is
exactly the $O(d\,m)$ sparse autoencoder \cite{cunningham2023, bricken2023}. Compression, projecting to a scalar,
and disambiguation, expanding to a sparse code, are not cheap and expensive
versions of one operation; they are different operations, and only the second
reads superposition. The gaps of Sec.~\ref{sec:dim}, $0.062$ on homogeneous
unknown material and $0.203$ on the heterogeneous mix, are direct measurements
of the truth information a single axis structurally cannot return.

The same limitation has a constructive side worth stating. One axis admits two
readings, not one. Its \emph{sign} classifies a statement:
$\mathrm{Re} > 0$ is the true side. Its \emph{separability}, how cleanly true
and false split on a given set of facts, measures how well the model knows
that material, because a model that does not know produces overlapping rather
than separated representations. The geometry that answers ``is this statement
true?'' therefore also answers, indirectly, ``does the model know this?''.
These are not two axes but two readings of one axis: the direction of the
arrow, and how sharp it is.

\textbf{Beyond single-direction probes on the residual stream}. All the
methods above, ours included, assume white-box access to the residual stream,
which is not always available and is not the only usable signal. Bar-Shalom et
al.~\cite{barshalom2025} treat the full activation tensor (layers by tokens)
as an image and train a vision-transformer detector on it, across several LLMs
simultaneously, with evaluation on a broad set of hallucination benchmarks.
Methods of this kind are more expressive than any single direction and are
trained for detection; our axis sits at the opposite end of the same
cost/expressivity curve, as the cheapest structural point, and the comparison
we draw is against that curve, not against any single method.

The consequence is a clear scope. The method's value is structural and cheap:
a training-free axis that on known facts rivals a trained probe at two dot
products per token; a quantitative relation between the readability of truth
and the model's knowledge of the fact; and a route to the polarity direction
that needs no polarity labels. It is not a substitute for sparse autoencoders
and is not presented as one.

\section{Summary of Part I}
Stripped of a decorative complex-plane framing, the positive results survive
on a real model under a falsifiable protocol: a training-free truth axis,
identified without labels up to one global sign, carries a permutation-tested
signal at cost $O(d)$; the gap between that axis and a full probe scales with
how well the model knows the fact, suggesting a knowledge-dependent
dimensionality of truth; and the difference SVD recovers the supervised
polarity direction of~\cite{burger2024} at cosine $0.959$ without polarity
labels. The one-dimensional reading withstands seven falsification
experiments, spanning a second dimension, depth-global structure, dynamics,
departments, energy, phase, and three-dimensional clustering, none of which
exhibits a richer separating coordinate. Beneath the axis, an exact
decomposition of the residual stream locates its origin: attention builds the
signal in the middle layers, causally, while the FFN erodes it after the
peak, and by the late layers the model has moved on from representing the
fact to predicting the next token. We read a result that resists seven
falsifications as strengthened, not weakened, interpret the residual overlap
on unknown facts as the visual signature of unresolved superposition, and
regard the axis as a light, honest, almost-free structural probe to sit
alongside generation.

\part{Mechanics and Geometry Across Families, and architectures}

\section{Setup}
\label{b:sec:setup}

14 models, 6 family: 2 instructed models, 2 Mixture-of-Experts models, and the other are base and dense models. All selected from diverse families, for evaluation the same phenomena in different architectures, and in models trained using different techniques and datasets.

\begin{table}[htbp]
\centering
\caption{Models. The truth peak $p$ is the block whose intact residual
maximizes held-out truth AUC (protocol of Part~I); ``block $b$''
denotes the output of transformer block $b$. Know-rate: fraction of $200$
CounterFact prompts whose true target the model reproduces under greedy
decoding (seed $0$), measured before any geometry. Axis and probe AUC:
CounterFact, $250$ pairs, held-out; and gap}
\label{b:tab:models}
\begin{tabular}{lccccccc}
\toprule
model & blocks & $d_{\mathrm{model}}$ & peak $p$ & know-rate & axis & probe & gap \\
\midrule
Qwen2.5-0.5b  & $24$ & $896$ & $13$ & $28.0\%$ ($55/200$) & $0.705$ & $0.762$ & $0.057$ \\
Qwen2.5-1.5b  & $28$ & $1536$ & $16$ & $32.0\%$ ($64/200$) & $0.792$ & $0.854$ & $0.062$ \\
Qwen2.5-3b    & $36$ & $2048$ & $17$ & $38\%$ ($76/200$) & $0.831$ & $0.882$ & $0.051$ \\
Qwen2.5-Coder-3b    & $36$ & $2048$ & $17$ & $32.0\%$ ($63/200$) & $0.790$ & $0.848$ & $0.058$ \\
Qwen2.5-instruct-3b    & $36$ & $2048$ & $17$ & $40.0\%$ ($79/200$) & $0.844$ & $0.883$ & $0.039$ \\
Llama-3.2-1B  & $16$ & $2048$ & $8$  & $36\%$ ($72/200$) & $0.690$ & $0.831$ & $0.141$ \\
Llama-3.2-3b  & $28$ & $3072$ & $10$  & $42.0\%$ ($85/200$) & $0.785$ & $0.903$ & $0.118$ \\
OLMo-2-1b  & $16$ & $2048$ & $9$  & $33\%$ ($66/200$) & $0.616$ & $0.644$ & $0.028$ \\
Gemma-2-2b    & $26$ & $2304$ & $12$ & $36.0\%$ ($71/200$) & $0.777$ & $0.875$ & $0.098$ \\
gemma-2-2b-it    & $26$ & $2304$ & $12$ & $34.0\%$ ($69/200$) & $0.876$ & $0.883$ & $0.007$ \\
tinyllama-1.1b  & $22$ & $2048$ & $13$ & $31.0\%$ ($62/200$) & $0.748$ & $0.831$ & $0.083$ \\
phi-2  & $32$ & $2560$ & $17$ & $38.0\%$ ($75/200$) & $0.694$ & $0.759$ & $0.065$ \\
granite-1b-a400m  & $24$ & $1024$ & $11$ & $26.0\%$ ($52/200$) & $0.662$ & $0.806$ & $0.144$ \\
granite-3b-a800m  & $32$ & $1536$ & $20$ & $34.0\%$ ($68/200$) & $0.682$ & $0.832$ & $0.150$ \\

\bottomrule
\end{tabular}
\end{table}

Two regularities and one boundary, extending the dimensionality claim
of Part~I: within each family the gap decreases as the know-rate
increases; across families a family-level offset remains, with Qwen
concentrating more of the truth signal on the single axis at comparable
knowledge. An earlier version of this table imported the Qwen2.5-1.5B row
from the first part's mixed-dataset column while the other rows were
measured on CounterFact alone, a protocol inhomogeneity caught by
re-measurement; all rows above come from one command under one protocol,
and the earlier phrasing that the gap ``tracks knowledge every time'' is
retired in favor of the two-part statement. The fifth row adds a third
family-level offset: at know-rate $36.0\%$ the Gemma gap ($0.098$) sits
between the Qwen offsets ($0.049$--$0.062$) and the Llama offsets
($0.83$--$0.141$ including tinyllama).

Dataset: \texttt{NeelNanda/counterfact-tracing} (pinned revision
\texttt{c945b08\ldots}, loaded as Parquet only), $250$ minimal pairs unless
stated. Decomposition runs use \texttt{float32} and are gated by the
additive identity check of Part~I (median relative errors
$0.65$--$1.23 \times 10^{-7}$ across the runs of this paper);
extraction-only runs use \texttt{bfloat16}. The spread of the four peaks in
relative depth ($0.32$--$0.54$) already anticipates the negative of
Sec.~\ref{b:sec:negatives}: no fixed depth, locates
truth, with the techniques used in this thesis.

\section{The anatomy of axis and truth in residual stream}
\label{sec:anatomy}

The first results established that truth is one-dimensional on
known facts. This section asks \emph{how the residual stream comes to carry
it}: which architectural component, attention or the FFN, produces, holds, or
degrades the signal, layer by layer. Same protocol and safeguards, but smaller margins in
places, and interpretations are marked as such. The decomposition and its
correctness gate are defined in Sec.~\ref{b:sec:methods}.

\subsection{Per-layer readability: a stable alternation}
\label{sec:alternation}

For each layer we fit the truth axis on $a_{\ell}$ and on $f_{\ell}$
separately and measure held-out AUC, repeating over $5$ cross-validation
seeds. A per-layer difference $\mathrm{AUC}(a_\ell) - \mathrm{AUC}(f_\ell)$
is counted as real only if its magnitude exceeds both $2\sigma$ (its own
across-seed standard deviation) and a floor of $0.03$. Table~\ref{tab:peak}
shows the peak block.

\begin{table}[htbp]
\centering
\caption{Truth AUC by component at the readability peak,
CounterFact, mean $\pm$ std over $5$ seeds. Residual = the summed stream
(reference).}
\label{tab:peak}
\begin{tabular}{lccc}
\toprule
 Model & residual & attention & FFN \\
\midrule
qwen-0.5b AUC & $0.700 \pm 0.004$ & $0.785 \pm 0.002$ & $0.742 \pm 0.003$ \\
qwen-1.5b AUC & $0.792 \pm 0.003$ & $0.845 \pm 0.003$ & $0.734 \pm 0.004$ \\
qwen-3b AUC & $0.829 \pm 0.003$ & $0.868 \pm 0.001$ & $0.848 \pm 0.002$ \\
qwen-Coder AUC & $0.792 \pm 0.004$ & $0.850 \pm 0.004$ & $0.827 \pm 0.001$ \\
qwen-3b-instruct AUC & $0.842 \pm 0.003$ & $0.865 \pm 0.002$ & $0.859 \pm 0.002$ \\
gemma-2 AUC & $0.776 \pm 0.002$ & $0.895 \pm 0.002$ & $0.815 \pm 0.002$ \\
gemma-2-it AUC & $0.876 \pm 0.003$ & $0.898 \pm 0.002$ & $0.882 \pm 0.002$ \\
llama-3.2-1b AUC & $0.690 \pm 0.004$ & $0.821 \pm 0.001$ & $0.772 \pm 0.004$ \\
llama-3.2-3b AUC & $0.787 \pm 0.001$ & $0.887 \pm 0.001$ & $0.814 \pm 0.001$ \\
tinyllama-1.1b AUC & $0.752 \pm 0.003$ & $0.849 \pm 0.002$ & $0.770 \pm 0.001$ \\
granite-a400m AUC & $0.657 \pm 0.003$ & $0.821 \pm 0.001$ & $0.788 \pm 0.002$ \\
granite-a800m AUC & $0.684 \pm 0.002$ & $0.822 \pm 0.002$ & $0.685 \pm 0.003$ \\
OLMo-2-1b AUC & $0.612 \pm 0.003$ & $0.674 \pm 0.003$ & $0.613 \pm 0.002$ \\
phi-2 AUC & $0.691 \pm 0.005$ & $0.822 \pm 0.002$ & $0.716 \pm 0.006$ \\

\bottomrule
\end{tabular}
\end{table}
The alternation survives re-seeding and reproduces structurally across both datasets and all evaluated model families. Across the entire population, a clear macro-phenomenon emerges: attention stably dominates readability throughout the early-to-intermediate layers, whereas the FFN contribution systematically catches up, matching or stabilizing its signal at deeper blocks. 

At each model's respective readability peak, attention consistently leads or closely balances the FFN contribution, showing a robust architectural regularity regardless of model scale or depth. \textbf{Interpretation (not yet causal):} attention appears to carry the truth signal most strongly through the middle band, including the peak block.

\section{Methods}
\label{b:sec:methods}

\subsection{Residual decomposition and its correctness gate}
\label{sec:decomp}

Within a block the residual stream is updated additively \cite{elhage2022},
\begin{equation}
h_{\ell+1} \;=\; h_{\ell} + a_{\ell} + f_{\ell},
\end{equation}
where $a_{\ell}$ is the attention contribution and $f_{\ell}$ the FFN
contribution, both captured by forward hooks. The residual stream studied
everywhere else in this paper is the sum; 
Every anatomy tool embeds an \textbf{identity check}: it verifies that
$a_{\ell} + f_{\ell}$ reconstructs the residual delta $h_{\ell+1} - h_{\ell}$.
The verdict uses the \emph{median} relative error across (sentence, layer)
rather than the maximum: where a block adds almost nothing the delta is near
zero and the ratio explodes, so the maximum is fragile while the median
reports the typical case. Observed median error: about $1.0 \times 10^{-7}$
on both datasets, so the decomposition is exact and the anatomy numbers are
trustworthy. (The single inflated layer, block $27$, has a near-zero residual
delta and is excluded from interpretation.) Anatomy runs use
\texttt{float32}: the identity fails spuriously in lower precision, because
subtracting large residual states to form small per-block deltas suffers
catastrophic cancellation.

\subsection{Frames and the fixed-frame contribution statistic}
\label{b:sec:fixedaxis}

\begin{definition}[Truth frame]
The \emph{post frame} $F_b^{\mathrm{post}}$ is the oriented truth axis $v_1$
fit (on training folds only, true side positive) on the block output
$h_{b+1} = h_b + a_b + f_b$; it contains $f_b$ and $a_b$. The \emph{pre
frame} $F_b^{\mathrm{pre}}$ is the same construction on the pre-FFN state
$h_b + a_b$; it contains $a_b$ but excludes $f_b$. A frame \emph{exists} at
$b$ when the axis fit there clears its permutation null held-out.
\end{definition}

The statistic for a contribution $c$ at layer $L$ measured against a frame
$F$ with axis $v_1$ is the held-out class gap and its effect size,
\begin{equation}
\mathrm{gap}(c; F) \;=\; \overline{(v_1 \cdot c)}\big|_{\mathrm{true}}
- \overline{(v_1 \cdot c)}\big|_{\mathrm{false}},
\qquad
d'(c; F) \;=\; \frac{\mathrm{gap}(c; F)}{\text{pooled std}}
\end{equation}
with the pooled standard deviation over the held-out projections of both
classes. A negative FFN gap means the FFN literally writes toward the false
side of the frame. Stability criterion, fixed in code before the
consolidation runs: a value is \emph{stable} if its magnitude exceeds both
$2\sigma$ across $5$ cross-validation seeds and a floor of $0.03$.

\subsection{The exact SwiGLU split}
\label{b:sec:swiglu}

The FFN computes $f = W_{\mathrm{down}}(g \odot u)$ with
$g = \mathrm{silu}(W_{\mathrm{gate}} x)$, $u = W_{\mathrm{up}} x$. The
projection of $f$ on $v_1$ is $w \cdot (g \odot u)$ with
$w = W_{\mathrm{down}}^{\top} v_1$, the residual axis pulled back through
the down-projection; no axis is ever fit in the expanded space, avoiding the
$p \gg n$ regime (more dimensions than samples) entirely. With the symmetric
intra-pair split ($\Delta g = g_t - g_f$, $\bar g = (g_t + g_f)/2$, likewise
for "$u$" the mixed term cancels and the identity is exact:
\begin{equation}
w \cdot (g_t \odot u_t - g_f \odot u_f)
= \underbrace{w \cdot (\Delta g \odot \bar u)}_{\text{gate term}}
+ \underbrace{w \cdot (\bar g \odot \Delta u)}_{\text{value term}}.
\label{b:eq:split}
\end{equation}
Verified numerically at $7.4 \times 10^{-8}$ relative error on real data.

\subsection{Sandwich normalization: the exact decomposition on Gemma-2}
\label{b:sec:sandwich}

Gemma-2 is not a pure pre-norm transformer: inside each block, the vector
added to the residual stream is not the module output but that output passed
through a further RMSNorm \\(\texttt{post\_attention\_layernorm} and
\texttt{post\_feedforward\_layernorm} in the reference implementation).
Three consequences, each verified. (i) \emph{Capture.} The additive
contributions $a_b, f_b$ are read by hooks at the outputs of the two
post-norms; hooking the modules themselves, correct on Qwen and Llama,
fails the additive identity check on Gemma at median relative error $0.94$,
and the corrected hooks restore it to $8.0 \times 10^{-8}$. Detection is by
the joint presence of the pre- and post-feedforward norms, never by the
name \texttt{post\_attention\_layernorm}, which exists on Llama and Qwen as
the \emph{pre}-norm of the MLP, a naming trap. (ii) \emph{Ablation.}
Replacing a module output by zero still contributes zero after the norm
(RMSNorm of the zero vector is zero), so every ablation and freeze of
Sec.~\ref{b:sec:interventions} carries over unchanged. (iii) \emph{The
split.} With $n(x) = (x/\mathrm{rms}(x)) \odot (1+\gamma)$ interposed
between the FFN and the residual, the projection of the true contribution
decomposes exactly into three terms,
\begin{equation}
v_1 \cdot \big(n(x_t) - n(x_f)\big)
= \bar{s}\,\underbrace{w \cdot (\Delta g \odot \bar u)}_{\text{gate}}
\;+\; \bar{s}\,\underbrace{w \cdot (\bar g \odot \Delta u)}_{\text{value}}
\;+\; \underbrace{\Delta s\, \bar P}_{\text{norm}},
\label{b:eq:sandwich}
\end{equation}
with $w = W_{\mathrm{down}}^{\top}\big((1+\gamma) \odot v_1\big)$,
$P = w \cdot (g \odot u)$, $s = 1/\mathrm{rms}(x)$,
$x = W_{\mathrm{down}}(g \odot u)$, and $\bar{\cdot}\,, \Delta$ the
intra-pair mean and difference. The identity is verified end-to-end
against the captured post-norm contribution at $5.0 \times 10^{-7}$; the
norm term measures class-driven rescaling by the post-FFN norm and is
reported alongside gate and value. Gemma's gating activation is a GELU
rather than a SiLU (GeGLU rather than SwiGLU); the $g \odot u$ structure
is identical, and we say \emph{gated-MLP split} where both are meant. On
the pre-norm models the code path is unchanged, the norm term is
identically zero, and a regression run on Qwen2.5-1.5B reproduces the
canonical numbers of Parts~I and II to the third decimal. \\

\subsection{Mixture of Expert}
\label{b:sec:MoE}

In a routed mixture-of-experts layer the FFN output is a weighted sum over
the experts the router selects for that token,
\begin{equation}
f_b(x) \;=\; \sum_{e \in \mathcal{T}_k(x)} g_e(x)\, E_e(x),
\qquad
E_e(x) \;=\; W^{e}_{\text{down}}\!\left(\sigma(W^{e}_{\text{gate}}x)
\odot W^{e}_{\text{up}}x\right),
\label{b:eq:moe}
\end{equation}
with $\mathcal{T}_k(x)$ the top-$k$ experts and $g_e$ their normalised router
weights. Two consequences follow, and they point in opposite directions.

The block-level decomposition is unaffected: $f_b$ enters the residual
additively exactly as in a dense layer, so $h_{b+1} = h_b + \lambda(a_b + f_b)$ holds
and the correctness gate passes ($1.2\times10^{-7}$ on both Granite models).
Everything in this Part that operates at block level therefore applies
unchanged.

The within-FFN split of Eq.~\ref{b:eq:split} does not survive. There is no
layer-level $g$ and $u$: each expert has its own, and $\mathcal{T}_k(x)$
varies per token, so the two members of a pair may be routed to different
experts and $\Delta g$, $\Delta u$ are not defined between them. The gate
versus value question has no layer-level counterpart here, and we report that
rather than constructing an analogue.

\subsection{Parallel blocks}
\label{b:sec:Parallel block}
The phi-2 architecture, that is also "non-gated" simply breaks down into a linear projection ("fc1" and "fc2"). Furthermore, this model features a variant of the "GELU" activation function known as NewGELU: $\text{FFN}(x) = \text{NewGELU}(x W_1 + b_1) W_2 + b_2.$

\subsection{Interventions}
\label{b:sec:interventions}

Ablating a component means replacing its additive contribution ($f_{\ell}$
or $a_{\ell}$) with the zero vector through a forward hook, for every layer
$\ell$ in a band (peak$+1$ to peak$+3$; readout at the band end, on every
model), during the forward pass: the weights are untouched, the module still
computes and its output is discarded, and the residual update reduces to
$h_{\ell+1} = h_{\ell} + a_{\ell}$ (for FFN ablation) or
$h_{\ell+1} = h_{\ell} + f_{\ell}$ (for attention ablation). Two
refinements. (i) A
\emph{frozen control} obtained for free: with both components zeroed over
the band, the readout state is identically the band-entry state, so the
frozen reference is read off the intact pass. (ii) \emph{Gate-freeze and
value-freeze}: at the reading position, $g$ (respectively $u$) is replaced
by its intra-pair mean cached from the intact run, killing the class-driven
variation of one stream while the FFN stays on.

\paragraph{On the validity of the freezes.} Any activation-level
intervention risks pushing the network off its data manifold, so three
properties of this design matter. The replacement value is the
\emph{intra-pair mean}: a convex combination of the two real activations
of two nearly identical inputs, which is as close to on-manifold as an
intervention of this kind gets. The two freezes are symmetric by
construction, so their contrast, controls for generic disruption: an off-manifold artifact has no
reason to respect the gate/value distinction. And the interventional
verdict agrees with the purely observational exact split of
Eq.~\ref{b:eq:split}, which touches nothing. 
We declared this risk.

\section{The peak-relative flip}

For each model the prediction was written before the run: under the
peak-fit post frame, the FFN's gap is pro-truth at the peak block $p$ and
flips to stable anti-truth at $p{+}1$.

\begin{table}[htbp]
\centering
\caption{The flip at peak$+1$ under the peak-fit post frame. CounterFact
$250$ pairs, $5$ seeds, permutation null with $100$ within-pair swaps
($\hat p = 0.0099$ on every row). All model rows from the uniform consolidation except for phi-2;}
\label{b:tab:flip}

\begin{tabular}{lcccc}
\toprule
model & $\mathrm{gap}_{\mathrm{ffn}}(p)$ & $\mathrm{gap}_{\mathrm{ffn}}(p{+}1)$ & $d'(p)$ & $d'(p{+}1)$ \\
\midrule
Qwen2.5-0.5B & $+0.200 \pm 0.002$ & $-0.110 \pm 0.002$ & $+0.63$ & $-0.64$ \\
Qwen2.5-1.5B & $+0.730 \pm 0.002$ & $-0.691 \pm 0.002$ & $+0.58$ & $-0.82$ \\
Qwen2.5-3B   & $+0.575 \pm 0.005$ & $-0.926 \pm 0.002$ & $+0.68$ & $-1.25$ \\
Qwen-Coder & $+0.647 \pm 0.003$ & $-0.685 \pm 0.003$ & $+0.82$ & $-1.26$ \\
Qwen-instruct & $+0.601 \pm 0.004$ & $-0.907 \pm 0.002$ & $+0.70$ & $-1.19$ \\
tinyllama-1.1b & $+0.054 \pm 0.001$ & $-0.080 \pm 0.001$ & $+0.35$ & $-0.88$ \\
Llama-3.2-1B & $+0.092 \pm 0.001$ & $-0.111 \pm 0.001$ & $+0.54$ & $-0.94$ \\
Llama-3.2-3B & $+0.087 \pm 0.001$ & $-0.167 \pm 0.001$ & $+0.49$ & $-1.26$ \\
gemma-2 & $+3.843 \pm 0.004$ & $-0.1673 \pm 0.007$ & $+0.94$ & $-0.59$ \\
gemma-2-instruct & $+9.868 \pm 0.009$ & $-1.846 \pm 0.006$ & $+1.39$ & $-0.44$ \\
granite-a800m & $+0.013 \pm 0.000$ & $-0.016 \pm 0.001$ & $+0.11$ & $-0.15$ \\
granite-a400m & $+0.043 \pm 0.000$ & $-0.025 \pm 0.000$ & $+0.69$ & $-0.55$ \\
OLMo-2-1b & $+0.021 \pm 0.001$ & $-0.074 \pm 0.001$ & $+0.14$ & $-0.51$ \\
phi-2 & $-0.429 \pm 0.002$ & $-0.422 \pm 0.000$ & $-0.84$ & $-0.71$ \\
\bottomrule
\end{tabular}
\end{table}

The peaks sit at different absolute and relative depths, the flip sits at peak$+1$ in all model. Aligned in
peak-relative coordinates and normalized by the flip amplitude, the
per-layer profiles collapse onto a common shape: rise into the peak, spike
at $+1$, decay toward zero. The measurement is
correct and reproduces; its interpretation, that the flip is anchored to
the peak, is what the next subsection tests. 
Note: We also report the failure of the phenomenon in Phi-2, since its architecture as previously stated does not feature pre and post processing within its blocks (parallel blocks).

\begin{figure}[h]
\centering

\includegraphics[width=0.75\textwidth]{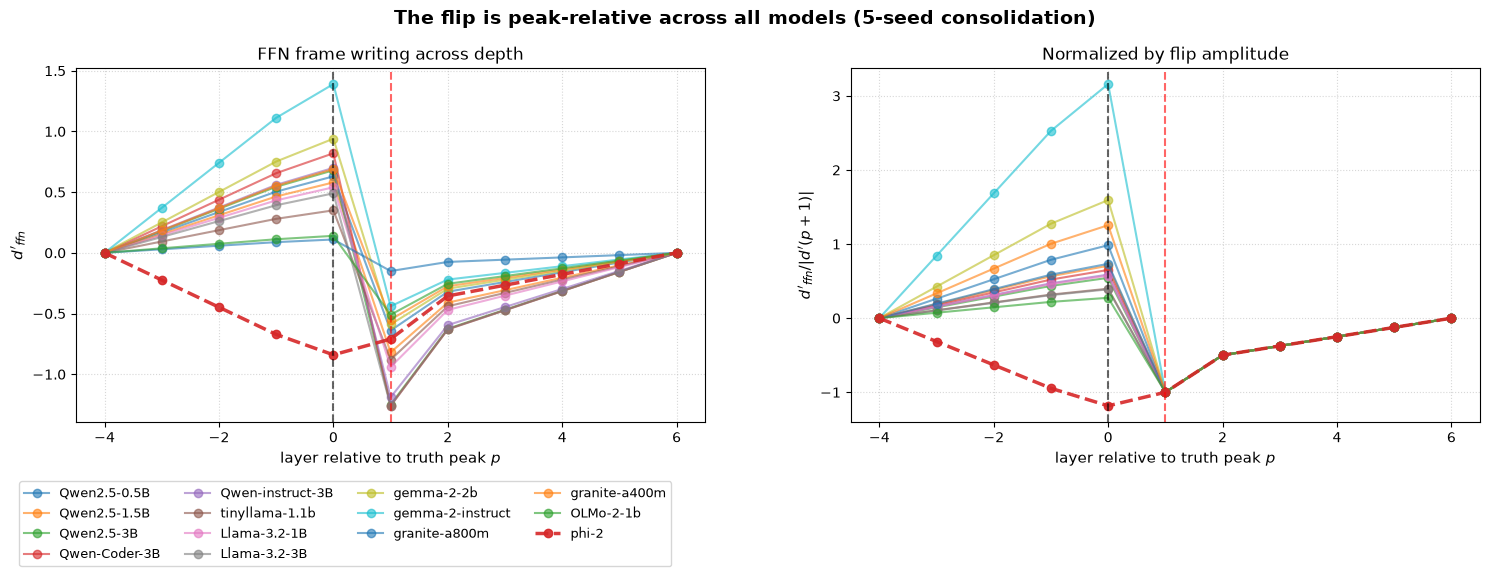}

\label{b:fig:asymmetry}
\end{figure}

\subsection{The axis-provenance control}
\label{b:sec:provenance}

The objection this method answers, raised by the author against the
author's own instrument, is that a result anchored to ``the block where the
ruler was fit, plus one'' could be a dynamic of the ruler rather than of
the FFN. The control: fit frames at several blocks
$b \in \{p{-}4, p{-}2, p, p{+}2\}$, both post and pre, and recompute the
full $d'$ profile of each component under each frame. The discrimination
criterion was fixed before the runs and validated on synthetic ground
truth: a planted global axis with a planted writer flip stays pinned under
every frame; a planted circular writer collapses under the pre frame
 while planted genuine alignment survives.
Under frames fit at $p{-}4$, $p{-}2$, $p$, $p{+}2$, the transition does not stay at $p{+}1$: it tracks the frame. On every dense-model the post-frame diagonal reads pro at $b$
and anti at $b{+}1$ for every $b$ where a frame exists
(Table \ref{tab:post}). The peak-anchored flip was the $b{=}p$ slice of a
relational pattern present at every such block. An accidental replication
is worth recording: a consolidation run launched with a wrong default
re-measured Qwen2.5-3B with the frame at block $15$, off its peak, and
returned the same pro/anti pair at $15/16$ ($+0.541$/$-0.545$, $5$ seeds,
$\hat p = 0.0099$); the accident confirms the phenomena at a frame not chose.

Under the pre frame the FFN's ``pro at own block'' term does not merely
shrink; it inverts sign (Table ~\ref{tab:pre}). Measured against the frame
that exists at the moment it writes, the FFN is anti-aligned. The
pro term of the post frame was, in substantial part, the frame measuring
its own ingredient: $f_b$ is inside $h_{b+1}$ and drags the axis fit there
toward itself. Where the axis is weak or absent, the pattern correctly
vanishes: frames fit before the Qwen-3B ignition (fit-block cosines to the
peak axis $0.049$ and $0.079$) and before the Llama-1B peak return noise.
What the failure bought: a general law instead of a special case, and a
live demonstration that the safeguards bind their owner.

\begin{table}[htbp]
  \centering

  \caption{Pre-frame diagonals: the same cells with $f_b$ excluded from the frame's data. The FFN inverts sign; attention, measured identically, is the mirror image. We will not show the results of all the models in these tables, as they exhibit the same phenomenon, and we will focus on the models that present architectural differences.}
  \label{tab:pre}
  \begin{tabular}{lccc}
  \toprule
  model (peak) & frame $b$ & $d'_{\mathrm{ffn}}(b)$ & $d'_{\mathrm{attn}}(b)$ \\
  \midrule
Qwen2.5-1.5B & $13$ / $15$ / $17$ & $-0.63$ / $-0.41$ / $-0.78$ & $+1.56$ / $+1.31$ / $+1.04$ \\
Qwen2.5-3B   & $14$ / $16$ / $18$ & $-0.14$ / $-1.16$ / $-0.46$ & $+0.26$ / $+1.38$ / $+1.40$ \\
Qwen2.5-Coder ($16$)   & $14$ / $16$ / $18$ & $+0.08$ / $-1.09$ / $-0.84$ & $-0.03$ / $+1.37$ / $+1.31$ \\
Qwen2.5-Instruct ($16$)   & $14$ / $16$ / $18$ & $-0.25$ / $-1.20$ / $-0.33$ & $+0.35$ / $+1.35$ / $+1.40$ \\
Llama-3.2-1B & $7$ / $9$          & $-1.17$ / $-0.82$           & $+1.26$ / $+0.82$ \\
Llama-3.2-3B & $7$ / $9$ / $11$   & $-0.81$ / $-1.43$ / $-1.41$ & $+0.99$ / $+1.63$ / $+1.75$ \\
phi-2 & $14$ / $16$ / $18$   & $-0.64$ / $-0.71$ / $-0.70$ & $+1.14$ / $+1.18$ / $+0.59$ \\
granite-3b-a800m & $17$ / $19$ / $21$   & $-0.12$ / $+0.47$ / $-0.05$ & $+0.17$ / $+1.12$ / $+0.22$ \\
granite-1b-a400m & $17$ / $19$ / $21$   & $+0.04$ / $-0.58$ / $-0.25$ & $+0.03$ / $+1.14$ / $+0.61$ \\
  \bottomrule
  \end{tabular}

  \vspace{1em}

  \caption{Post-frame diagonals: per fit block $b$, the FFN's held-out $d'$ at $b$ and at $b+1$. Pro at $b$, anti at $b+1$, wherever a frame exists.}
  \label{tab:post}
  \begin{tabular}{lccc}
  \toprule
  model (peak) & frame $b$ & $d'_{\mathrm{ffn}}(b)$ & $d'_{\mathrm{ffn}}(b{+}1)$ \\
  \midrule
Qwen2.5-1.5B ($15$) & $13$ / $15$ / $17$ & $+0.61$ / $+0.57$ / $+0.52$ & $-0.79$ / $-0.82$ / $-0.73$ \\
Qwen2.5-3B ($16$)   & $14$ / $16$ / $18$ & $+0.06$ / $+0.69$ / $+0.73$ & $-0.09$ / $-1.23$ / $-0.79$ \\
Qwen2.5-Coder ($16$)   & $14$ / $16$ / $18$ & $-0.01$ / $+0.83$ / $+0.61$ & $+0.02$ / $-1.24$ / $-0.86$ \\
Qwen2.5-Instruct ($16$)   & $14$ / $16$ / $18$ & $+0.09$ / $+0.70$ / $+0.84$ & $-0.12$ / $-1.17$ / $-0.72$ \\
Llama-3.2-1B ($7$)  & $5$ / $7$ / $9$    & $+0.13$ / $+0.54$ / $+0.19$ & $-0.07$ / $-0.95$ / $-0.54$ \\
Llama-3.2-3B ($9$)  & $7$ / $9$ / $11$   & $+0.37$ / $+0.49$ / $+0.57$ & $-0.42$ / $-1.25$ / $-1.39$ \\
phi-2 & $14$ / $16$ / $18$   & $-0.08$ / $-0.06$ / $+0.00$ & $-0.42$ / $-0.84$ / $-0.63$ \\
granite-3b-a800m & $17$ / $19$ / $21$   & $+0.06$ / $+0.47$ / $+0.20$ & $+0.27$ / $+0.11$ / $+0.07$ \\
granite-1b-a400m & $17$ / $19$ / $21$   & $+0.07$ / $+0.68$ / $+0.25$ & $-0.04$ / $-0.54$ / $-0.49$ \\
  \bottomrule
  \end{tabular}

\end{table}

\begin{law}[Propagate vs.\ overwrite]
\label{b:law:relational}
Let $F$ be a truth frame that exists at block $b$, and let $L \geq b$ be a
layer whose contributions are not contained in $F$ (any $L$ for pre frames at
$b = L$; any $L > b$ for both frame types). Then, across the dense models
studied, $d'(f_L; F) < 0$ and $d'(a_L; F) > 0$ wherever both are stable:
attention propagates frames it did not write, and the FFN opposes them.
\end{law}

The decisive zone is the sub-diagonal of the pre matrices: contributions at
layer $L$ measured against frames of blocks strictly before $L$, which contain
neither $f_L$ nor $a_L$ and are clean of every containment effect. There
(Fig.~\ref{b:fig:asymmetry}) the FFN is negative and attention positive nearly
everywhere a frame exists. At the peak, under the same clean frame, the two
components are near mirror images: $+1.31 / -0.41$ (Qwen-1.5B),
$+1.38 / -1.16$ (Qwen-3B), $+1.26 / -1.17$ (Llama-1B), $+1.63 / -1.43$
(Llama-3B).

The law dissolves the earlier implicit picture of an axis being eroded like an
object. The axis is not an object in the model but a moving statistical
summary of where the class difference lives, and the FFN's class-signed writes
are what successive frames are made of. This is the dynamical face of the key-value write-head physiology 
attributed to MLPs~\cite{geva2021}, and a candidate microscopic writer 
for the macroscopic findings of~\cite{meng2022} and the rotational 
dynamics reported in~\cite{rotational2026}.

The scope clause is not decorative. Sec.~\ref{b:sec:phimoe} reports where the
regularity holds and where it does not, and the exception is architectural
rather than accidental.

\begin{figure}[h]
\centering

\includegraphics[width=0.55\textwidth]{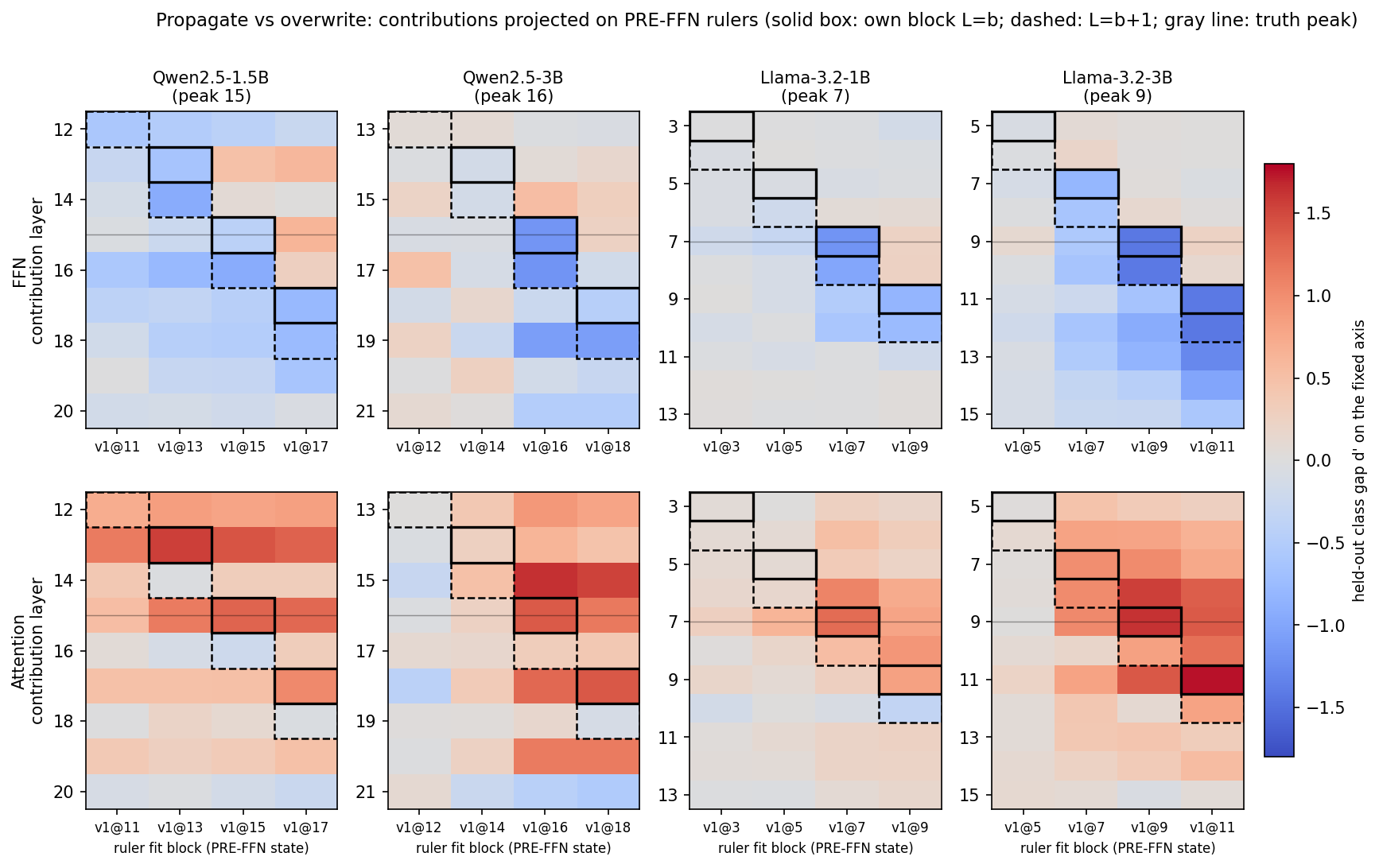}
\caption{Propagate vs.\ overwrite: held-out $d'$ of each
component's contribution (FFN above, attention below) against pre frames
fit at four blocks. The FFN contribution is negative wherever a frame exists; the
attention contribution is positive.}
\label{b:fig:asymmetry}
\end{figure}

\subsection{Two registered predictions on non-sequential architectures}
\label{b:sec:phimoe}

Two architectures in the set do not follow the sequential of dense, and we
registered a prediction for each before the runs. Both failed, in different
directions, and the failures are informative.

\paragraph{Phi-2: the law holds, the pro-truth phase does not.}
We predicted that Phi-2 would not exhibit the relational pattern, on the
assumption that parallel-block architectures lack the distinct ``pre'' and
``post'' dynamics of sequential models: attention and the FFN read the same
input, so $h_b + a_b$ is a quantity one can compute but not a state the model
passes through.

Under the pre frame the prediction is falsified. Phi-2 shows the mirror image
like every dense sequential model: $-0.64 / -0.71 / -0.70$ for the FFN against
$+1.14 / +1.18 / +0.59$ for attention (Table~\ref{tab:pre}). The relational
law is therefore not a consequence of the sequential ordering of modules
within a block. It concerns the geometry of the contributions relative to the
accumulated state, and it holds even where the frame is a construction rather
than a state.

Under the post frame something else is true, and it is the observation the
figure caption records. The pro-truth term at the block's own position, which
is $+0.5$ to $+0.8$ on the dense sequential models, is absent on Phi-2:
$-0.08 / -0.06 / +0.00$ (Table~\ref{tab:post}). The flip table says the same
in another coordinate: Phi-2 is the only model in the set whose $d'(p)$ is
negative ($-0.84$), so there is no pro-truth phase at the peak from which to
turn away.

The two statements are not in tension. The relational law under a clean frame
survives on parallel blocks; the pro-truth phase at the peak under the
frame's own output does not. We report the first as a falsified prediction and
the second as an architectural difference, and we treat both as leads rather
than as settled mechanism.

\paragraph{Mixture of experts: the deviation is confined to the routed
component.}
We predicted that the two Granite models would align with the dense models.
They do not, and the way they fail is specific.

At the peak frame, attention behaves exactly as everywhere else: $+1.12$ on
Granite-3b-a800m and $+1.14$ on Granite-1b-a400m, against a range of $+1.14$
to $+1.63$ across the seven dense models. It is the FFN that deviates:
$+0.47$ and $-0.58$, against $-0.41$ to $-1.43$ on the dense models, with the
sign inverted on the larger of the two.

Attention is dense in both Granite models; the FFN is not. The deviation is
therefore confined to the component the router controls, which is what
Eq.~\ref{b:eq:moe} would predict. In a routed layer the block's write is a
sparse sum over the experts selected for that token, and the two members of a
pair may be routed to different experts: the pair difference then mixes two
different functions rather than measuring one function on two inputs, and any
class-signed statistic built on it is attenuated.

This is a mechanism we can name and did not test. The direct check is the
expert overlap between the two members of a pair, which the instrumentation
already exposes; if the overlap is low where the attenuation is large, the
reading is confirmed. We record it as a lead and as the first item of the
future work in Sec.~\ref{sec:limits}.

\section{The erosion: real, frame-free, and causally carried by the value
stream}
\label{sec:erosion}

This section, separated two phenomena that the
peak-anchored reading conflated: frame-relative rotation, which is what the
flip measures, and frame-free decay of class signal after the peak, which is
the erosion proper. This section establishes three things about the second, in
order: that it is real, that it is causal rather than a projection artifact,
and what inside the FFN carries it.

\subsection{The tug-of-war}

Over the band from peak$+1$ to peak$+3$ we compare four conditions at the band
end: intact, with the FFN zeroed across the band, with attention zeroed, and a
frozen reference. The frozen reference costs no extra pass: with both modules
zeroed over the band the state at the readout is identically the state entering
the band, so it is read off the intact run at the band entrance. It answers a
question the other three do not: what readability would be if the band did
nothing at all.

We report two protocols. In the first the axis is refit inside each condition, on the held-in pairs of each fold and read on the held-out ones; in the second the axis is the intact one, fit before any intervention. The first asks whether anything is still readable, the second whether the original direction survives. They are different questions and they do not always agree. 

\begin{table}[htbp]
\centering
\begin{tabular}{lrrrr}
\hline
model & frozen & intact & FFN-off & attn-off \\
\hline
Qwen2.5-1.5B    & 0.793 & 0.712 & 0.815 & 0.708 \\
Qwen2.5-3B      & 0.831 & 0.793 & 0.846 & 0.738 \\
Llama-3.2-1B    & 0.691 & 0.551 & 0.744 & 0.531 \\
Llama-3.2-3B    & 0.785 & 0.740 & 0.847 & 0.593 \\
TinyLlama-1.1B  & 0.748 & 0.574 & 0.785 & 0.536 \\
OLMo-2-1B       & 0.616 & 0.555 & 0.664 & 0.527 \\
Phi-2           & 0.694 & 0.655 & 0.711 & 0.630 \\
Granite-1b-a400m& 0.662 & 0.607 & 0.682 & 0.562 \\
Granite-3b-a800m& 0.682 & 0.570 & 0.701 & 0.581 \\
\hline
\end{tabular}
\caption{Causal ablation over the band, held-out refit AUC at the band end.
FFN-off restores signal on every model in the set; attention-off is at or below
intact everywhere. The frozen column is the intact state entering the band.}
\end{table}

FFN-off exceeds intact on all nine models, by $+0.019$ (Qwen2.5-3B) to $+0.211$
(TinyLlama-1.1B). Attention-off is at or below intact everywhere except
Granite-3b, where it is $+0.011$, inside the sampling error of $0.022$ at $250$
pairs.

The two protocols separate on Qwen2.5-1.5B: FFN-off reaches $0.815$ when the
axis is refit, but $0.757$ when read on the intact readout axis. Removing the
FFN recovers separability, and does so partly along a direction that is not the
original one. The recovery is therefore not simply the removal of an eraser.

\subsection{Rotation or erosion: the causal test}

The per-layer table shows the FFN's mean pair difference turning against the
axis after the peak. That is an observation about direction, not a
demonstration of effect, and the objection raised in session was precise: a
negative gap could be a rotating write direction seen through a fixed axis
rather than genuine anti-truth writing.

The two hypotheses separate algebraically before any measurement. Write the
FFN's contribution as $f = f_\parallel + f_\perp$ with
$f_\parallel = (f\cdot v)v$. A component orthogonal to the axis cannot change
the sign of the projection, since $(x+y)\cdot x = \|x\|^2$ for $y \perp x$;
only an anti-parallel component can. Held-out AUC on a \emph{fixed} axis is
likewise blind to orthogonal additions. So the two hypotheses make opposite
predictions under a directional intervention.

We therefore replace the FFN's write in a band after the peak with one of its
projections and read held-out AUC downstream on the intact axis, fitted on
training folds before any intervention. A control removes a random direction of
the same construction instead of the truth axis, which separates the effect of
the direction from the effect of removing something.

Across $18$ block-interventions in six models (Qwen2.5-3B, Gemma-2-2b,
OLMo-2-1B, Phi-2, Granite-3b-a800m, TinyLlama-1.1B), removing the parallel
component recovers readability, mean $+0.014$, sign test
$p = 7.3\times10^{-5}$. The random-direction control gives a mean of $-0.0001$
with a maximum absolute value of $0.001$.

The erosion is therefore real and causal. It is also small: about one
hundredth of AUC per block, which does not account for most of the post-peak
decay. We report the magnitude rather than only the sign.

\paragraph{The orthogonal write is not inert.}
At the flip block, removing the \emph{orthogonal} component instead
\emph{damages} downstream readability: $-0.021$ on Qwen2.5-3B, $-0.033$ on
OLMo-2-1B, $-0.030$ on TinyLlama-1.1B, the three models with the sharpest
flips. On Granite-3b, Gemma-2 and Phi-2, where the flip is milder, the same
removal is neutral or slightly positive.

An orthogonal component cannot affect the projection at its own level. It can
only matter if later blocks read it and convert it back into on-axis signal.
The FFN's post-peak write is therefore not purely an eroder: part of it is
material that downstream blocks use. This refines rather than contradicts the
tug-of-war picture: attention propagates, the FFN overwrites, and what it
writes is not noise but the substrate of later frames.

\subsection{Inside the SwiGLU: the value stream is the carrier}

The FFN's contribution factorises as $W_{\text{down}}(g \odot u)$ with
$g = \sigma(W_{\text{gate}}x)$ and $u = W_{\text{up}}x$. The two factors have
different roles: the gate decides how much passes, the value decides what
passes. We separate them causally by replacing one stream's last-token
activation with its intra-pair mean, cached from the intact run, over the band.
This kills that stream's class-driven variation while leaving its scale and the
other stream untouched. The FFN stays on.

The choice of hook matters and is not free: the gate is read at the output of
the activation, not of the projection. Since $\sigma$ is non-linear,
$\sigma(\text{mean})$ is not $\text{mean}(\sigma)$, and freezing before the
activation reverses the sign of the result.

\begin{table}[htbp]
\centering
\begin{tabular}{lrrrrr}
\hline
model & (readout) & frozen & gate-freeze & value-freeze  \\
\hline
Qwen2.5-1.5B & 0.712 & 0.793 & $+0.040$ & $+0.089$ \\
Qwen2.5-3B   & 0.793 & 0.831 & $+0.008$ & $+0.064$ \\
Llama-3.2-1B & 0.551 & 0.691 & $+0.028$ & $+0.123$  \\
Llama-3.2-3B & 0.740 & 0.785 & $+0.056$ & $+0.122$  \\
OLMo-2-1B    & 0.555 & 0.616 & $+0.045$ & $+0.114$  \\
Gemma-2-2b &0.765 &0.715 & $+0.032$ & $+0.068$ \\
TinyLlama-1.1B & 0.574& 0.748& $+0.113$ & $+0.103$  \\
\hline
\end{tabular}
\caption{Causal freezes over the band, held-out refit AUC, change relative to
intact. Frozen is the intact state entering the band, i.e.\ what readability
would be if the band did nothing. Silencing the value stream recovers signal on
every model, and by a factor of two to eight more than silencing the gate.}
\end{table}

The split does not apply to every architecture in the set. Phi-2 uses an
ungated MLP with $\texttt{fc1}/\texttt{fc2}$ and a GELU non-linearity, so there
is no gate to freeze. The two Granite models use a routed mixture-of-experts
FFN: each expert is itself gated, but there is no layer-level gate, and which
experts are active varies per token. On both families the block-level
decomposition holds exactly (identity check $1.2\times10^{-7}$ on Granite);
it is the within-FFN split that has no counterpart. We report this.

The earlier hypothesis that the gate carries the erosion, that which
neurons open switches the stream toward next-token context, remain falsified in the
same direction on every model where the split applies. In the following tests, only TinyLlama exhibits a significant change resulting from the alteration of the gating component; however, it remains 1 model with a Transformer architecture identical to Llama, differing only in the pre-training dataset.

\paragraph{What the negative does and does not license.}
We do not propose a replacement variable. Two considerations argue against
reading further structure into this column at the present sample size. The
sampling error on a single held-out AUC at $250$ pairs is about $0.022$, so
values between $0.038$ and $0.055$ sit at one to two standard errors from the
floor; only Llama-3.2-3B is comfortably outside the noise. And the indicator is
a difference of two AUCs computed on the same pairs, so its own error is
smaller than that of either term but was not estimated here.

The honest statement is therefore narrower than a negative result about scale:
with $250$ pairs and this floor, the discrimination is not testable. Reporting
it as decided in either direction would overstate what the design supports.

\section{A pre-registered negative, cross-family: the static crossover does not track the peak}
\label{b:sec:negatives}

Motivated by an external observation that value-to-query weight norms scale
with depth~\cite{headgenome}, the gauge-invariant version was tested: does
the layer where the writing circuit overtakes the routing circuit predict
the truth peak? The prediction for the Llama models, registered before
their runs, was that it does not. Outcome (Table~\ref{b:tab:crossover}): the
crossover sits at a stable late relative depth on all four models
($0.61$--$0.75$) while the peak wanders ($0.32$--$0.54$); the two coincide
only on Qwen2.5-1.5B, the accident that spawned the hypothesis the first
working note falsified across scale. The depth-scaling of the circuits may
be real and general; it is not where truth lives. The late-layer OV
explosion (Qwen-1.5B sums $338.9$/$322.0$ at layers $26$--$27$) instead
matches the decoupling of Table \ref{b:tab:models} (peak registered).

\subsection{Static circuit norms}
\label{b:sec:circuits}

Per attention head (GQA-aware), the gauge-invariant circuit norms
$\lVert W_q^{\top} W_k \rVert_F$ (routing) and $\lVert W_o W_v \rVert_F$
(writing)~\cite{headgenome}. Individual weight norms are gauge-dependent
($W_q \to \alpha W_q$, $W_k \to W_k / \alpha$ leaves the function
unchanged); the products are not. A gauge self-test gates the computation.
The \emph{crossover} is the first layer from which the median-normalized
OV/QK ratio stays above $1$ for at least three consecutive layers, a
criterion fixed in code before any run.

\begin{table}[htbp]
\centering
\caption{The OV/QK crossover (first layer sustaining ratio $>1$ for three
consecutive layers) against the truth landmarks, four models. Distance is
to the nearer of peak and flip.}
\label{b:tab:crossover}
\begin{tabular}{lcccc}
\toprule
model & crossover & peak / flip & distance & rel.\ depth (cross vs.\ peak) \\
\midrule
Qwen2.5-1.5B & $17$ & $15$ / $16$ & $1$ & $0.61$ vs.\ $0.54$ \\
Qwen2.5-3B   & $24$ & $16$ / $17$ & $7$ & $0.67$ vs.\ $0.44$ \\
Llama-3.2-1B & $12$ & $7$ / $8$   & $4$ & $0.75$ vs.\ $0.44$ \\
Llama-3.2-3B & $17$ & $9$ / $10$  & $7$ & $0.61$ vs.\ $0.32$ \\
\bottomrule
\end{tabular}
\end{table}

\paragraph{A methodological note on automatic verdicts.} During the
regeneration of this table, the circuit tool's hardcoded landmark
annotations (written for one model) mislabeled the truth peak on the other
three and printed a spurious coincidence verdict for Llama-3.2-3B. It was
the third automatic verdict to misread its own numbers in this project's
history, after two in the provenance tooling. All released tools now print
matrices and distances with no verdict prose; The raw norm tables, which depend only on the weights, were
unaffected.

\part{The arrangement of relational categories}
\label{part:arrangement}

\section{The extended model set}
\label{sec:arr-models}

The set spans fourteen models, six families and five architectural families:
pre-norm (Qwen2.5, Llama-3.2, TinyLlama, Granite), sandwich-norm (Gemma-2),
post-norm (OLMo-2), parallel blocks (Phi-2), and a routed mixture-of-experts
FFN (Granite-3.1).

Ranked by mean corrected agreement against all other models:
Qwen2.5-Coder-3B $0.860$, Qwen2.5-3B-Instruct $0.857$, Qwen2.5-3B $0.855$,
Phi-2 $0.830$, TinyLlama-1.1B $0.817$, Granite-1b-a400m $0.815$,
OLMo-2-1B $0.804$, Qwen2.5-0.5B $0.799$, Llama-3.2-3B $0.789$,
Gemma-2-2b $0.779$, Granite-3b-a800m $0.774$, Qwen2.5-1.5B $0.762$,
Llama-3.2-1B $0.761$, Gemma-2-2b-it $0.670$.

The two mixture-of-experts models sit inside the range rather than apart from
it, so a sparse routed FFN does not displace the arrangement.

\section{Methods}
\label{sec:arr-methods}

\subsection{Sentence construction}
\label{sec:arr-convention}

Every state is read at the last token of a sentence built as prompt, space,
target, followed by a period.  As previously reported and explained in Part I, \ref{sec:convention}

Dictionaries built under the two conventions are not interchangeable, and the
bridge between them is measured. The global axes differ at cosine $0.474$,
more than sixty degrees apart, while the arrangement survives at Mantel
$+0.775$. Rebuilding the sentences without the period and refitting recovers
an archived axis at cosine $1.000$, confirming that the two pipelines read the
same position and differ in nothing else.

The arrangement is therefore invariant under a choice that rotates the axis
almost completely. We report this as a result rather than as a caveat.

\subsection{Reliability, attenuation ceiling, disjoint samples}
\label{sec:arr-reliability}

An agreement between two matrices cannot be read without knowing how well each
is reproducible. We therefore define three quantities and report them together
throughout.

\emph{Reliability} of a model is the mean Mantel correlation between
dictionaries bundles built for that model under identical settings, differing only in
the sampling seed. With four seeds this is a mean over six within-model pairs,
and we report its spread.

\emph{Ceiling} for a pair of models is $\sqrt{r_A r_B}$, the largest agreement
two imperfect measures of one common factor can display. If one measure repeats
at $0.80$ and the other at $0.90$, they cannot agree beyond $0.85$ even when
the same geometry underlies both.

\emph{Corrected} agreement is observed divided by ceiling. It is always
reported next to the raw value: dividing by a low ceiling amplifies noise, and
a corrected value above one indicates the model is violated rather than an
agreement better than possible.

Matched seeds mean matched sentences, so two models compared at the same seed
share their sampling noise. We therefore take the \emph{cross-seed} mean as the
primary statistic and the matched-seed mean only as a control. The inflation
from a shared sample is small and measured: $+0.035$ (standard error $0.016$)
between Gemma-2-2b and Llama-3.2-3B, the two most reliable models in the set.

\subsection{Per-category axes, transfer, decoding}
\label{sec:arr-percat}

CounterFact is grouped by Wikidata relation; the top-$K$ relations by
unique pairs are used ($K{=}8$ for the registered falsification, $60$ pairs
each): P103 mother tongue, P1412 languages spoken, P176 manufacturer, P27
citizenship, P30 continent, P37 official language, P413 sport position,
P495 country of origin etc. Per-category truth axes are fit at the peak block,
each oriented within its own category, so the sign of a cross-category
cosine is information: negative means the shared direction reads the other
category's truth backwards. Four measurements: (i) signed cosine matrices
at the peak and at an early block (block $2$, the lexical surface control);
(ii) the held-out \emph{transfer matrix}, cell $(i,j)$ being the held-out
AUC of category $i$'s axis on category $j$'s pairs; (iii) nearest-centroid
\emph{decoding} of the category from the FFN's class-signed write direction
$\Delta f$ at peak$+1$, against the same decoder on early residual
differences (the lexical baseline); (iv) the \emph{Mantel test}, a
permutation test for the correlation between two distance matrices, here
applied under category-label permutation to compare the two families
signed-cosine arrangements.

\subsection{The sign gauge}
\label{sec:arr-gauge-def}

Each per-category axis is identified up to sign (Lemma~\ref{lem:labels}), so
the sign of a cross-category cosine is not an observable until an orientation
is declared. Orienting each axis within its own category, by the sign of that
category's own class margin, is the obvious choice and the fragile one: the
margin is thin exactly where the model knows little, and a thin margin is a
coin. A reference that does not depend on any single category is needed.

\paragraph{Definition.}
Let $\mathcal{C} = \{c_1,\dots,c_K\}$ be the categories and let
$v_1,\dots,v_K \in \mathcal{S}^{d-1}$ be their unit axes, each fixed by an
arbitrary initial orientation. Write $C \in \mathbb{R}^{K\times K}$ for the
matrix of signed cosines, $C_{ij} = v_i \cdot v_j$, and let $u$ be a leading
eigenvector of $C$, that is
\begin{equation}
C\,u \;=\; \lambda_1 u,
\qquad
\lambda_1 \;\geq\; \lambda_2 \;\geq\; \dots \;\geq\; \lambda_K .
\label{eq:gauge-eig}
\end{equation}
The \emph{spectral consensus gauge} declares the orientation of category $k$ to
be $\operatorname{sign}(u_k)$, with $\operatorname{sign}(0) = +1$ by
convention, and replaces $C$ with
\begin{equation}
\widetilde{C} \;=\; S\,C\,S,
\qquad
S \;=\; \operatorname{Diag}\!\big(\operatorname{sign}(u)\big)
\;\in\; \{-1,+1\}^{K\times K}.
\label{eq:gauge-apply}
\end{equation}
Equivalently, each axis is replaced by $\operatorname{sign}(u_k)\,v_k$. The
gauge uses no pooled axis and no external reference: it reads the orientation
that the categories agree on among themselves. It is idempotent, since applying
it to $\widetilde{C}$ leaves the matrix unchanged, and $u$ is itself defined up
to a global sign, which we fix by requiring a positive majority.

\paragraph{The gauge cannot manufacture structure.}
Only relative signs are observable. For any assignment $S$ of the form above,
the product of cosines around a closed cycle is invariant:
\begin{equation}
\widetilde{C}_{ij}\,\widetilde{C}_{jk}\,\widetilde{C}_{ki}
\;=\;
s_i s_j \cdot s_j s_k \cdot s_k s_i \; C_{ij} C_{jk} C_{ki}
\;=\;
C_{ij} C_{jk} C_{ki},
\label{eq:gauge-cycle}
\end{equation}

We report the \emph{frustration} of a bundle as the fraction of triangles
whose cycle product is negative,
\begin{equation}
\phi \;=\; \frac{1}{\binom{K}{3}}
\left|\left\{ (i,j,k) : C_{ij}C_{jk}C_{ik} < 0 \right\}\right| ,
\label{eq:frustration}
\end{equation}
which by Eq.~\ref{eq:gauge-cycle} is invariant under every sign assignment and
is therefore a property of the geometry rather than of the orientation chosen.

since $s_j^2 = 1$ for every $j$. A set of axes whose cycles are frustrated,
that is whose cycle products are negative, admits no assignment that makes all
pairwise cosines positive. The gauge therefore cannot create agreement where
none exists; it can only remove the arbitrariness of the initial orientation
and leave whatever obstruction the geometry contains. Sec.~\ref{sec:arr-gauge}
reports one such obstruction that survives the gauge on every model.

\paragraph{When the sign is not identifiable.}
Nothing in Eq.~\ref{eq:gauge-eig} guarantees that $u$ is well determined. When
$\lambda_1$ and $\lambda_2$ are close, the leading eigenvector rotates under
small perturbations of $C$ and the signs it declares are not stable. We
therefore report the relative eigengap
\begin{equation}
\gamma \;=\; \frac{\lambda_1 - \lambda_2}{\lambda_1}
\label{eq:gauge-gap}
\end{equation}
alongside every gauged bundle, as an \emph{a priori} condition on
identifiability: it is computed from $C$ alone, before any sign is assigned and
without reference to any result. Categories whose margin $|u_k|$ falls below a
declared threshold are reported unsigned rather than oriented, and the tooling
marks them in the bundle metadata.

\begin{table}[H] 
\centering
\small 
\caption{Relative eigen-gap $\gamma$ Eq.\ref{eq:gauge-gap} and frustation $\phi$ Eq.\ref{eq:frustration}, sd is over seeds.}
\label{b:tab:frustation}
\begin{tabular}{l c c c c c} 
\toprule
Model & $K$ & Seeds & $\gamma$ & sd & $\phi$ \\ 
\midrule
Llama-3.2-1B & 33 & 4 & $0.740$ & $0.024$ & $0.135$ \\
Llama-3.2-3B & 33 & 4 & $0.779$ & $0.026$ & $0.143$ \\
OLMo-2-1B & 33 & 4 & $0.643$ & $0.041$ & $0.305$ \\
Qwen2.5-0.5B & 33 & 4 & $0.634$ & $0.029$ & $0.279$ \\
Qwen2.5-1.5B & 33 & 4 & $0.817$ & $0.016$ & $0.190$ \\
Qwen2.5-3B & 33 & 4 & $0.777$ & $0.020$ & $0.267$ \\
Qwen2.5-3B-Instruct & 33 & 4 & $0.795$ & $0.013$ & $0.208$ \\
Qwen2.5-Coder-3B & 33 & 4 & $0.722$ & $0.019$ & $0.282$ \\
gemma-2-2b & 33 & 4 & $0.854$ & $0.009$ & $0.050$ \\
gemma-2-2b-it & 33 & 4 & $0.918$ & $0.001$ & $0.017$ \\
granite-1b-a400m & 33 & 4 & $0.659$ & $0.031$ & $0.296$ \\
granite-3b-a800m & 33 & 4 & $0.649$ & $0.048$ & $0.248$ \\
phi-2 & 33 & 4 & $0.711$ & $0.013$ & $0.244$ \\
tinyllama-1.1b & 33 & 4 & $0.720$ & $0.046$ & $0.219$ \\
\bottomrule
\end{tabular}
\end{table}

\section{Reliability}
\label{sec:arr-rel}

At $K=33$ relations and $n=60$ pairs per relation, restricted to categories
clearing the knowledge gate, reliabilities range from $0.764$ to $0.917$ across
fourteen models. On the full category set they range from $0.586$ to $0.871$.

The distance between the two is itself a measurement, and Sec. \ref{sec:arr-gate}
gives it a reading: categories the model does not know contribute rows
estimated on noise, which are not reproducible and drag the full-set
coefficient down without saying anything about geometry.

\begin{table}[htbp]
\centering
\caption{Reliability at K = 33, n = 60: Mantel correlation between dictionaries of the same model, built with four msampling seed. Full uses all the 33/34 categories of counterfact; restricted use keeps all the categories with AUC above 0.6 (Knowledge gate on both). The $p$ column is the lowest of the six pairs at the permutation of 1/1001.}
\label{tab:reliability_results}
\begin{tabular}{l c c c c c c}
\toprule
\textbf{Model} & \textbf{Pairs} & \textbf{Full} & \textbf{Restricted} & \textbf{Early} & \textbf{$p$-value} & \textbf{Worst $K'$} \\
\midrule
Llama-3.2-1B       & 6 & $0.812 \pm 0.050$ & $0.892 \pm 0.043$ & $ 0.059 \pm 0.068$ & 0.0010 & $K'=22$ \\
Llama-3.2-3B       & 6 & $0.759 \pm 0.100$ & $0.797 \pm 0.072$ & $-0.040 \pm 0.038$ & 0.0010 & $K'=24$ \\
OLMo-2-1B          & 6 & $0.586 \pm 0.052$ & $0.905 \pm 0.007$ & $ 0.088 \pm 0.065$ & 0.0010 & $K'=13$ \\
Qwen2.5-0.5B       & 6 & $0.744 \pm 0.030$ & $0.868 \pm 0.030$ & $ 0.017 \pm 0.104$ & 0.0010 & $K'=19$ \\
Qwen2.5-1.5B       & 6 & $0.871 \pm 0.024$ & $0.877 \pm 0.033$ & $-0.013 \pm 0.049$ & 0.0010 & $K'=24$ \\
Qwen2.5-3B         & 6 & $0.856 \pm 0.017$ & $0.912 \pm 0.021$ & $ 0.073 \pm 0.099$ & 0.0010 & $K'=24$ \\
Qwen2.5-3B-Instruct& 6 & $0.857 \pm 0.027$ & $0.917 \pm 0.030$ & $ 0.034 \pm 0.076$ & 0.0010 & $K'=24$ \\
Qwen2.5-Coder-3B   & 6 & $0.753 \pm 0.029$ & $0.877 \pm 0.036$ & $ 0.116 \pm 0.097$ & 0.0010 & $K'=20$ \\
gemma-2-2b         & 6 & $0.870 \pm 0.034$ & $0.875 \pm 0.038$ & $ 0.090 \pm 0.173$ & 0.0010 & $K'=27$ \\
gemma-2-2b-it      & 6 & $0.757 \pm 0.097$ & $0.764 \pm 0.100$ & $ 0.038 \pm 0.129$ & 0.0020 & $K'=29$ \\
granite-1b-a400m   & 6 & $0.669 \pm 0.036$ & $0.885 \pm 0.044$ & $ 0.031 \pm 0.044$ & 0.0010 & $K'=17$ \\
granite-3b-a800m   & 6 & $0.709 \pm 0.053$ & $0.870 \pm 0.033$ & $ 0.038 \pm 0.100$ & 0.0010 & $K'=20$ \\
phi-2              & 6 & $0.869 \pm 0.030$ & $0.894 \pm 0.028$ & $ 0.018 \pm 0.045$ & 0.0010 & $K'=19$ \\
tinyllama-1.1b     & 6 & $0.756 \pm 0.031$ & $0.849 \pm 0.070$ & $ 0.083 \pm 0.151$ & 0.0010 & $K'=22$ \\
\bottomrule
\end{tabular}
\end{table}

\begin{figure}[htbp]
\centering
\includegraphics[width=\textwidth]{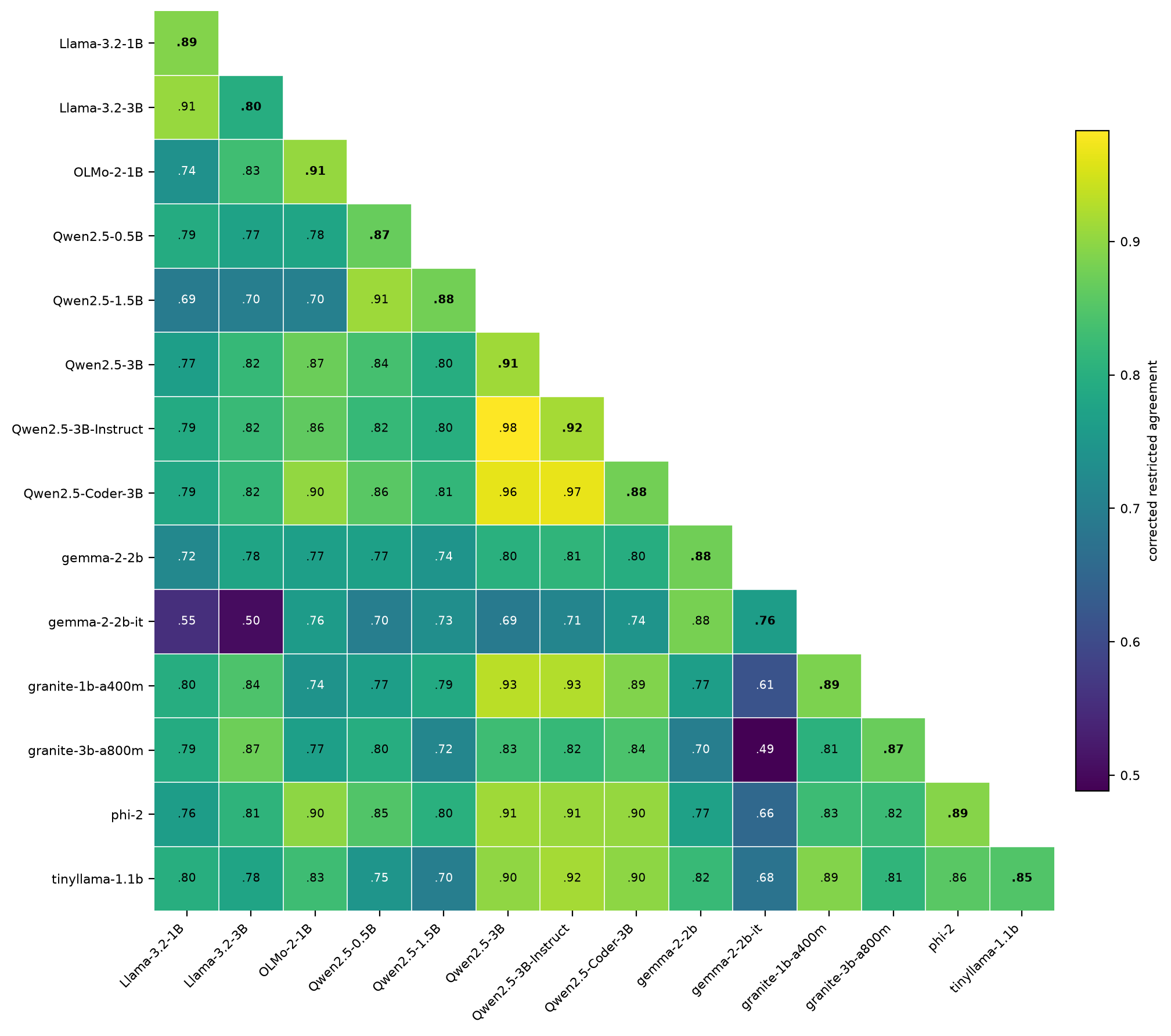}
\caption{Pairwise arrangement agreement across the fourteen models, ordered by
family. Off-diagonal cells give the agreement corrected for attenuation on the
restricted category set; the diagonal gives each model's own restricted
reliability across four sampling seeds, which is the ceiling its row cannot
exceed. Values run from 0.49 to 0.98 and the colour scale is set to that range,
not to $[0,1]$.}
\label{fig:arrangement-matrix}
\end{figure}

\section{The law, on disjoint samples}
\label{sec:arr-law}

Across fourteen models, six families and five architectures, all $91$ pairwise
agreements are positive. Corrected for attenuation on the restricted set the
mean is $0.798$, with a minimum of $0.488$ and a maximum of $0.983$. Exactly
one of the $91$ pairs fails to reach $p<0.05$.

\section{A family effect}
\label{sec:arr-family}

Within a family the corrected agreement averages $0.861$ over $15$ pairs;
across families it averages $0.786$ over $76$ pairs. The difference is $+0.076$
with a standard error of $0.020$, that is $3.8$ standard errors.

The arrangement is therefore largely shared across families \cite{platonic2024} with a measurable
family-specific component, rather than identical across families. The earlier
formulation, which asserted the stronger claim, is retracted in favour of this
one.

\section{Surface control, down to layer 0}
\label{sec:arr-surface}

The objection this section answers is that the shared arrangement could be
lexical structure of the evaluation material rather than geometry built by the
models. We answer it at two depths.

At the early block, agreement across the $55$ cross-model pairs ranges from
$-0.026$ to $+0.112$, mean $+0.027$; none exceeds $0.15$. The margin between
peak and early ranges from $+0.310$ to $+0.772$, mean $+0.554$.

The stronger form is at layer $0$, the embeddings, before any block has
computed anything. There all $528$ off-diagonal cosines equal $1.0000$ with a
standard deviation of zero: the per-category axes are the same vector, and no
category structure exists at all. The arrangement is built by the blocks; it is
not inherited from the embedding matrix.

\section{The knowledge gate}
\label{sec:arr-gate}

Where a model does not know a fact there is no truth direction, so that
category's axis is estimated on noise and its row in the matrix is not
reproducible. Restricting to categories whose within-category AUC clears $0.6$
on both models is therefore not data cleaning: it is the Part~I prediction
applied to Part~II.

The effect is large and asymmetric. On OLMo-2-1B, with a know-rate of $34\%$,
only $13$ of $33$ categories clear the gate; full reliability is $0.586$ while
restricted reliability is $0.905$, the highest of the fourteen. A model that
knows few facts and knows them well is not an unreliable measurement: it is a
measurement with a smaller domain.

The number of surviving categories orders the models by how much they know:
$K'=13$ (OLMo-2-1B), $17$ (Granite-1b-a400m), $22$ (TinyLlama-1.1B), $24$
(Qwen2.5-3B), $27$ (Gemma-2-2b).


\section{The sign gauge and its identifiability}
\label{sec:arr-gauge}

\paragraph{Provenance of this section.}
The bundles analysed here predate the sentence convention adopted in the rest
of this work: they were built without the final period,
pairs per relation. Their coefficients are therefore not comparable with those
of Sec.~\ref{sec:arr-law}, and the bridge between the two conventions is
measured in Sec.~\ref{sec:arr-convention}: the global axes differ at cosine
$0.474$, while the arrangement survives at Mantel $+0.775$. We keep the section
because the pathology it documents is real and its diagnosis is what motivated
the gauge; we mark the protocol so that no number here is read against a number
there.

The instability described below is a failure of \emph{measurement} under three
conditions that the final protocol removes. Eight categories give $28$
off-diagonal cells, too few to estimate a correlation stably. A sample of $888$
pairs per relation dilutes each category with facts the model does not know,
and the singular spectrum of short sentences flattens accordingly. And no sign
convention was declared, so each category's orientation was fixed by whatever
its own margin happened to be. Read in that light, what follows is a diagnosis.

\paragraph{The instability: per-category orientation is a thin gauge.}
Each axis is identified by Lemma up to sign (Part I), and the sign
was fixed within the category. We observed this phenomenon using bundles produced under the convention that omits the "." suffix; therefore, this section directly concerns the canonical analyses predating this version (we also publish bundles from earlier versions in the dedicated repository). We present this result as an observation of informational value. That orientation margin is thin exactly
where the model knows little, and a thin margin is a coin. The seed study
exposed the coin: re-sampling the $60$ Qwen pairs under six seeds left the
unsigned structure $|C|$ essentially unchanged but produced sign storms in
half the runs (three of six signed matrices anti-correlated with the
canonical one, Mantel $-0.42$/$-0.26$/$-0.21$; sign repair restores
$+0.46$/$+0.59$/$+0.73$, identifying coordinated flips of two to three
axes). The same pathology governs density: at $888$ pairs the full
cross-family comparison collapses to $+0.233$ ($p = 0.15$), every category
subset containing P495 fails while the same test excluding P495 returns
$+0.621$ ($p = 0.005$), and the repair diagnostic identifies P495 as the
flipped axis. The low know-rates of the affected relations place the
instability where the dimensionality relation of Part~I predicts weak,
diffuse signal.

\paragraph{A failed cure, reported: the pooled global axis is capturable.}
The first repair we registered was the global-axis gauge. The control that
killed it is worth the space it takes: on the Qwen $K{=}8$ pool the stored
global direction has cosine $0.99$ with the axis of P30 (continent), the
one category that anti-aligns with the language block. The dominant
direction of a heterogeneous pool is a property of the mixture: whichever
category contributes the largest, most coherent difference vectors
captures it, so the referee inherits exactly the problem it was appointed
to judge. On Llama, whose pool is dominated by a compact high-knowledge
block, the same estimator is healthy; the failure is knowledge-dependent,
like everything else in this series.

\paragraph{What the final protocol does to the pathology.}
The three conditions above are removed together in the campaign of
Sec.~\ref{sec:arr-law}: $33$ categories instead of $8$, giving $528$
off-diagonal cells; $60$ pairs per relation instead of $888$, which keeps each
category on the facts the model is most likely to know; and the spectral
consensus gauge applied by declaration rather than by repair.

The instability does not survive it. Restricted reliabilities across the
fourteen models range from $0.764$ to $0.917$, against the $0.53$ that the
$K=8$ pool returned under the same statistic. Sign storms of the kind reported
above, three signed matrices in six anti-correlated with their own
counterparts, have no counterpart in the extended campaign.

We state this as a methodological result rather than as a rescue. The pathology
was not worked around; the conditions that produced it were identified and
removed, and each removal is declared. That the diagnosis predicted which
condition mattered, and that removing them worked, is the reason the gauge is reported here.

\section{The measured dictionary}
\label{sec:arr-dictionary}

The arrangement of Sec.~\ref{sec:arr-law} is not only a set of coefficients: it
is an artefact, and the artefact is small. For each model,
\texttt{campagna.py --stages dizionario} writes one loadable bundle containing
the $K$ per-category axes with their declared orientation, the pooled global
axis, the FFN write centroids, the signed cosine matrix at the peak and at the
early block, the transfer matrix, and the full protocol metadata: model, peak
block, pairs per relation, sentence suffix and seed. At $K=33$ and
$d \leq 3072$ this is a $[K \times d]$ object of a few hundred kilobytes.

What the bundle supports is a $K$-direction monitor: given a state, one inner
product per category returns how far it lies along each category's truth
direction, at a total cost of $O(Kd)$ and with no training. Its baseline
performance is quantified above, and its provenance is declared in the same
file, so two bundles can be compared only when the five protocol fields match.

It is not a substitute for dictionary learning, and the reason is structural
rather than a matter of engineering effort. A projection is an inner product:
it returns how much of a chosen direction is present, not which features are.
Reading superposition requires one question per dictionary atom, which over an
overcomplete dictionary of size $m$ is exactly the $O(dm)$ sparse autoencoder.
The argument is given in Sec.~\ref{sec:discussion}, and the probe gaps measured
in Part~I, $0.062$ on homogeneous unknown material and $0.203$ on the
heterogeneous mix, are direct measurements of the truth information a
projection structurally cannot return.

The measured dictionary is therefore offered as a baseline with a declared
domain: cheap, training-free, and quantified on the material of this study. A
sparse autoencoder trained on the same models would be the natural comparison,
and the form that comparison should take is stated in
Sec.~\ref{sec:limits}.

\section{A pre-registered negative: lineage}
\label{sec:arr-lineage}

Two families in the set are distilled by documented account: Meta reports that
Llama-3.2 1B and 3B were obtained by structured pruning from Llama-3.1 followed
by knowledge distillation, and Google reports training Gemma-2 2B and 9B with
the predictions of a larger teacher. We pre-registered the hypothesis that the
shared arrangement reflects shared lineage, and tested it with two models
trained from scratch without a teacher: OLMo-2-1B on Dolma and TinyLlama-1.1B
on SlimPajama.

The hypothesis is not supported. OLMo-2-1B ranks seventh of fourteen and
TinyLlama-1.1B fifth, both above Gemma-2-2b and above both Llama models.

The TinyLlama comparison is the cleaner of the two because it holds
architecture and size fixed and varies only provenance. TinyLlama-1.1B and
Llama-3.2-1B share the Llama architecture at $1.1$ and $1.24$ billion
parameters and reach a full-probe ceiling of $0.831$ each, to three decimals.
Distillation from a $70$-billion-parameter teacher leaves no measurable trace
on this property.

\section*{Limitations and future work}
\label{sec:limits}

\begin{itemize}
\item All tests were conducted on small-scale models, demonstrating the phenomena on models with up to 3 billion parameters due to high computational costs with our strict FP32 precision requirement. Extending the evaluation to larger models is a limitation of the present work and a direction for future work. 
\item The know-rate is a string-match lower bound under greedy decoding and is
  not comparable across families; no claim here compares know-rates between
  families.
\item Attenuation correction assumes a single common factor with independent
  errors. Disjoint seeds are what makes the independence assumption defensible;
  matched-seed comparisons violate it, which is why they are reported only as
  a control.
\item Gemma-2-2b-it is a persistent outlier: mean corrected agreement $0.670$,
  the lowest of the fourteen, and the only pair in the campaign failing
  $p<0.05$. It agrees with its own base at $0.882$, so it is not a broken
  measurement; instruction tuning displaced its arrangement relative to every
  other model, while the same intervention on Qwen2.5-3B moves almost nothing
  ($0.983$ between base and instruct). We report the dissociation without
  explaining it.
\item Reliability is itself an estimate with its own error, and that error
  propagates into every corrected value. With six pairs per model the spread we
  report alongside each mean is the quantity to read.
\item Everything is tested using the CounterFact dataset, a dataset constructed for patching at the subject token rather than for reading a verdict at the final token. This is the limitation that affects every part, and we acknowledge it.
\item Based on our empirical evaluations, with the falsification of the prediction on Mixture of expert architecture, this unexpected divergence prompted us to plan a future study regarding the anatomy of truth representation in MoEs specifically, within the routing mechanism and its experts.
\item Furthermore, we plan to extend this geometric analysis from the comprehensive residual stream down to the level of individual attention heads. Preliminary exploratory data suggests that comparing the arrangement maps (cosine similarity matrices) of the total residual stream against those derived from specific attention heads could unveil how localized attention patterns aggregate into the "macro-phenomenon of propagating truth". Although this specific line of inquiry is currently supported only by early, limited evidence, formalizing this multi-scale structural comparison remains a primary objective for our future work.
\item A sparse autoencoder trained on these models would be the natural
comparison for the measured dictionary of Sec.~\ref{sec:arr-dictionary}, and
we would consider it justified against matched nulls if it aligns with the
per-category axes, respects the semantic arrangement, reproduces the declared
signs including the inverted loading on the continent relation, and improves on
the measured dictionary at class-signal reconstruction and category decoding.
Either outcome is content: a dictionary that fails these tests is not a better
instrument, and one that passes them explains what the projection compresses.

\end{itemize}

\section*{Conclusion}
\label{sec:conclusion}

One instrument was applied to three questions. The axis is training-free, it
is identified without labels up to a single global sign, and it costs one
inner product per token; everything reported here is what that instrument can
and cannot see.

\paragraph{What holds.}
The truth signal on a known fact is essentially one-dimensional: a single
direction obtained from paired differences reaches AUC $0.938$ on curated
facts, and no second coordinate, no depth trajectory and no polar
reformulation adds more than $+0.02$. What the axis misses is not a second
truth direction but the truth of facts the model does not know: the gap to a
full probe grows from $0.051$ on curated pairs to $0.062$ on CounterFact and
to $0.203$ on heterogeneous material. Dimensionality is knowledge-dependent
and category-dependent, and both statements are measured rather than asserted.

The mechanics are consistent across fourteen models, six families and five
architectural placements. Attention leads the FFN at the readability peak on
every model; the FFN's contribution turns against the axis one block later;
and under a frame that contains neither contribution, attention is positive
and the FFN negative wherever both are stable. A directional intervention
separates the two candidate mechanisms: removing the component parallel to the
axis recovers readability in $17$ of $18$ block-interventions, while removing
a random direction of the same construction changes nothing. Inside the FFN
the anti-truth write is carried by the value stream, not by the gate.

The axis is a mixture, and the mixture is arranged. Per-category axes agree
across models in all $91$ pairwise comparisons, with a mean of $0.798$ once
each model's own measured reliability is divided out. The agreement is not
lexical: at the initial block it is near zero on every pair, and at the
embedding layer the arrangement does not exist at all, since the $528$
off-diagonal cosines are identical to four decimals. What the blocks build is
therefore built by the blocks.

\paragraph{What was withdrawn.}
Three readings from earlier versions of this work did not survive contact with
their own controls. The complex-plane framing, in which falsehood appeared as
a phase rotation, reduces to the sign of the real axis in polar coordinates and
is retracted. The stability floor of $0.03$ was applied to a raw projection
whose scale varies by a factor of thirty-five across models, and is replaced by
a dimensionless criterion reported alongside it. And the pro-truth term at a
block's own position, under a frame fit on the block's own output, was in
substantial part the ruler measuring its own ingredient; under a clean frame
it inverts sign.

Two pre-registered predictions failed. The shared arrangement does not reflect
shared distillation lineage: two models trained from scratch without a teacher
rank above both distilled families, and the cleanest comparison, TinyLlama-1.1B
against Llama-3.2-1B at matched architecture and size, reaches the same
full-probe ceiling of $0.831$ to three decimals. And the tug-of-war between
attention and the FFN after the peak, recorded as a trait of scale on four
models, is not ordered by parameter count once ten are measured; at the present
sample size the discrimination is not testable in either direction.

\paragraph{Scope.}
All models here are below four billion parameters, and the identity gates that
license every decomposition require float32, which is why they are. Every
measurement rests on CounterFact, a resource built for activation patching at
the subject token rather than for reading a verdict at the final one; the
sentence construction that adapts it is declared, and its consequences are
measured rather than assumed.

\paragraph{A registered prediction.}
Every category measured here is a retrieval relation: the model either has the
fact stored or it does not. Whether the arrangement extends to a category whose
truth is \emph{computed} rather than retrieved is open, and it is the question
we consider most worth asking next.

The material is arithmetic minimal pairs in which the \emph{operator} varies
while arguments and result are held fixed, for example ``$2+2=4$'' against
``$2-2=4$'', in symbolic and in verbal form. Varying the operator rather than
the result matters: changing the result turns the task into completion of a
memorised string, whereas changing the operator holds the surface almost
constant and varies the operation. The operator sits mid-sentence, so under the
period convention the state is read at a position where the operator's identity
can only have arrived by propagation.

The prediction is that an arithmetic category, once it clears the knowledge
gate, takes a position in the arrangement rather than sitting orthogonal to it,
and that its agreement with retrieval categories is positive but below the
within-retrieval mean of $0.798$. Two outcomes would falsify it: an arithmetic
axis orthogonal to the retrieval arrangement would show that the shared
geometry is specific to stored facts; an agreement at or above the retrieval
mean would show that the arrangement is indifferent to how truth is obtained,
which we do not expect.

The feasibility check comes first, and it is the principle of
Sec.~\ref{sec:why} applied to ourselves: the know-rate of arithmetic
completions must be measured before any geometry, because a model that does
not compute produces noise that reads as distributed structure. On models of
this size we expect single-digit arithmetic to clear the gate and little else,
which would bound the study rather than prevent it.

\paragraph{}
The discipline that produced these results is the one we would defend
independently of them. Predictions were written before the runs. Correctness
gates halt the pipeline rather than warn, and one of them refused an entire
architecture until a scalar in its configuration was accounted for. Where a
control bit its owner, the correction is in the text as a result. A claim that
has survived its author's own best attack is worth more than one that has never
been attacked, and several claims here did not survive.

\section*{Acknowledgments}

The author thanks Boaz Nadler for detailed and constructive comments on an
earlier version of Part~I, which prompted, among many improvements, the
formalization of its Method section and Lemma~\ref{lem:labels}; the
standards of definition and precision of those comments are applied
throughout this manuscript.

\section*{Reproducibility}
\label{sec:repro}

Two codebases produced the numbers in this manuscript, and they are released
separately because they answer to different constraints.

The \emph{canonical} scripts are those of the original series and produce
Part~I in full, together with the four-model tables of Part~II. They target
pre-norm and sandwich-norm architectures and are released unmodified, so that
every published number remains reproducible by the code that produced it.

The \emph{library} was written to extend the same measurements to
architectures the canonical scripts do not cover: post-norm, parallel blocks,
and routed mixture-of-experts layers. It reimplements the same geometry, and
where both run they agree: the global axis of a canonical dictionary is
recovered at cosine $1.000$ when the sentence convention matches, and the
frames and ablation tables reproduce the canonical values on Qwen2.5-1.5B.
That agreement is the reason the two are reported together rather than as
separate studies.
\\
\texttt{https://github.com/Francesco-Marhel/TruthProbe (repository for canonical scripts)}
\\
\texttt{pip install git+https://github.com/Francesco-Marhel/truthprobe-core.git}
(library, commit \texttt{3317b62})
\\
\begin{table}[htbp]
\centering
\caption{The command that produced each table. Every run is deterministic
given its seed; the dataset revision is pinned in each script and echoed in
the log header.}
\label{tab:repro}
\small
\begin{tabular}{ll}
\toprule
Table & Command \\
\midrule
Part~I, all & \texttt{truth\_probe.py} \texttt{}{(flags in \texttt{REPRODUCING.md})} \\

Know-rates            & \texttt{campagna.py --stages behav} \\
Per-layer anatomy     & \texttt{campagna.py --stages anatomy --seeds 5 --perm 100} \\
Peak-relative flip    & \texttt{campagna.py --stages flip} \\
Pre/post frames       & \texttt{campagna.py --stages frames} \\
Causal ablation       & \texttt{campagna.py --stages ablazione} \\
Directional test      & \texttt{causale.py --modo direzionale --per-blocco} \\
Gate/value freeze     & \texttt{gatevalue.py} \\
Dictionaries          & \texttt{campagna.py --stages categories,dizionario,gauge} \\
Part~III, all         & \texttt{riepilogo\_mantel.py <dir> --perms 999 --soglia 0.6} \\
Depth curve           & \texttt{mantel\_per\_livello.py --a <b1> --b <b2>} \\
\bottomrule
\end{tabular}
\end{table}

Every model path, peak block and protocol parameter appears in the log header
of each run, and the dictionaries are released with their metadata: model,
block, number of pairs, sentence suffix and seed. Two bundles are comparable
only when those five fields match, and the tooling refuses to compare them
otherwise.

\end{document}